\documentclass[10pt,journal,compsoc]{IEEEtran}
\ifCLASSOPTIONcompsoc
  \usepackage[nocompress]{cite}
\else
  \usepackage{cite}
\fi
\ifCLASSINFOpdf
   \usepackage[pdftex]{graphicx}
\else
\fi

\usepackage{lineno,hyperref}
\modulolinenumbers[5]
\usepackage[bottom]{footmisc}
\usepackage{breqn}
\usepackage{amssymb}
\usepackage{float}
\usepackage{stfloats}
\usepackage{xcolor}

\usepackage{multirow}
\begin{document}
\title{DE-GAN: A Conditional Generative Adversarial Network for Document Enhancement}

\author{Mohamed Ali~Souibgui
        and~Yousri~Kessentini% <-this % stops a space
\IEEEcompsocitemizethanks{\IEEEcompsocthanksitem M. A. Souibgui is with   Computer Vision Center, Edifici O,  Bellaterra, 08193, Spain and Computer Science Department, Universitat Autònoma de Barcelona, Spain.\protect\\
E-mail: \emph{msouibgui@cvc.uab.es}
\IEEEcompsocthanksitem Y. Kessentini, Digital Research Center of Sfax, B.P. 275, Sakiet Ezzit, 3021 Sfax, Tunisia,  MIRACL Laboratory, University of Sfax, Sfax, Tunisia.  \protect\\
E-mail: yousri.kessentini@crns.rnrt.tn

}% <-this % stops an unwanted space
}

\IEEEtitleabstractindextext{%
\begin{abstract}
Documents often exhibit various forms of degradation, which make it hard to be read and substantially deteriorate the performance of an OCR system.  In this paper, we propose an effective end-to-end framework named Document Enhancement Generative Adversarial Networks (DE-GAN) that uses the conditional GANs (cGANs) to restore severely degraded document images. To the best of our knowledge, this practice has not been studied within the context of generative adversarial deep networks. We demonstrate that, in different tasks (document clean up, binarization, deblurring and watermark removal), DE-GAN  can produce an enhanced version of the degraded document with a high quality.
In addition, our approach provides consistent improvements compared to state-of-the-art methods over the widely used DIBCO 2013, DIBCO 2017 and H-DIBCO 2018 datasets, proving its ability to restore a degraded document image to its ideal condition. The obtained results on a wide variety of degradation reveal the flexibility of the proposed model to be exploited in other document enhancement problems.
\end{abstract}

}

\maketitle

\IEEEdisplaynontitleabstractindextext

% \fi
\IEEEpeerreviewmaketitle

\IEEEraisesectionheading{\section{Introduction}\label{sec:introduction}}

\begin{figure*}[ht]
\centering
\includegraphics[scale=0.45]{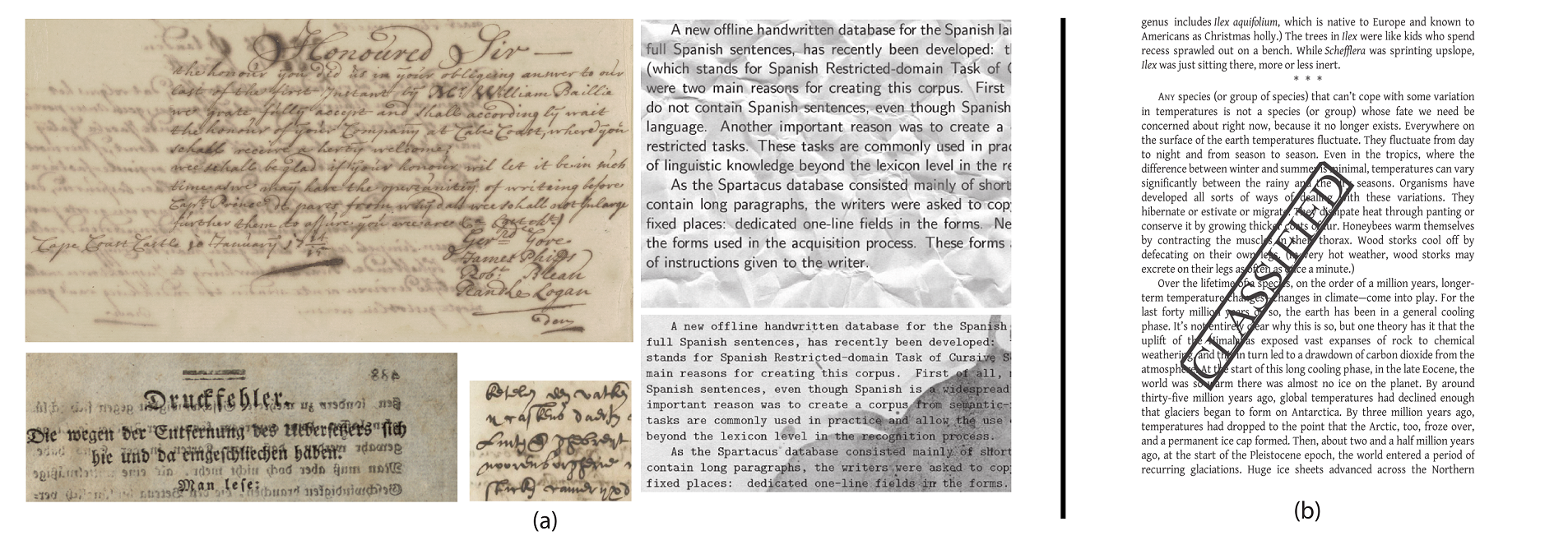}
\caption{Examples of the documents used in this study: (a): Degraded documents, (b): A document with dense watermark.} \label{fig1}
\end{figure*}

\IEEEPARstart{A}{utomatic} document processing consists in the transformation into a form  that is comprehensible by a computer vision system or by  a human. Thanks to the development of several public databases, document processing has made a great progress in  recent years \cite{art31,art32}. However, this processing is not always effective when  documents are degraded.  Lot of damages  could be done to a document paper. For example: 
Wrinkles, dust, coffee or food stains, faded sun spots and lot of real-life scenarios \cite{art19}.  Degradation could be  presented also in the scanned documents because of the bad conditions of digitization like using the smart-phones cameras (shadow \cite{art30}, blur \cite{art37}, variation of light conditions, distortion, etc.).  Moreover, some documents could contain watermarks, stamps or annotations. The recovery is even harder when certain types of these later take the text place for instance in cases where the stains color is the same or darker than the document font color  (Fig. \ref{fig1} shows some examples).  Hence, an approach to recover a clean version of the degraded document  is needed.

In this study, we are focusing on two document enhancement problems. Degraded documents recovery, i.e., to produce a clean (grayscale or binary) version of the document given any type of degradation, and watermark removal.  The faced obstacles are as follows: Overlaps of noise or watermarks with the text, dense watermarks, intense dirt or degradation can cover the entire text and reading it becomes very hard, there is no prior knowledge about the degradation or the watermark that should be removed, etc.  An ideal system should be good in performing two tasks simultaneously, removing the noise and the watermarks as well as retaining the text quality in the document images.    

Recently, a great success is made by   deep neural networks in natural images generation and restoration, especially deep convolutional neural networks (auto-encoders and variational auto-encoders (VAE)) \cite{art1,art46,art66} and generative adversarial networks (GANs) \cite{art2,art18}.  GANs, which were   introduced in  \cite{art13}, are now considered as the ideal solution for  image generation problems \cite{art18} with a high quality, stability and variation compared to the auto-encoders. Generative models gained more attention because of the ability of  capturing high dimensional probability distributions, imputation of missing data and dealing with multimodal outputs. Despite that, document analysis research community is not benefiting enough from those approaches, yet. Using them  is very limited, for instance, in font translation \cite{art55}, handwritten profiling \cite{art33} and staff-line removal from a music score images \cite{art34}, where a promising results were found. 

 In \cite{art2}, Isola et al. show that  conditional generative adversarial networks (cGANs), a variation of GANs, performs good in image-to-image translation (labels to facade, day to night, edges to photo, BW to color, etc.).  While GANs learn a generative model of data, conditional GANs (cGANs) learn a conditional generative model, where it conditions on an input image and generate a corresponding output image. Since document enhancement follows the same process, means, we want to preserve the text and remove the damage in a conditioned image, cGANs shall be the suitable solution, and this is what motivated us for this study.  

The main contributions of this paper are: As primary, to the best of our knowledge, this is the first occurrence of GANs, conditional GANs specifically,  in  a framework that addresses different document enhancement problems (clean up, binarization and watermark removal). 
Second, we used a simple but flexible architecture  that could be exploited  to tackle any document degradation problem. Third, we introduce a new document enhancement problem  consists in dense watermarks and stamps removal. Finally, we experimentally prove that our approach achieves a higher performance compared to the state-of-the-art methods in degraded document binarization.

The rest of the paper is organized as follows. In Section 2, a summary of previous works on document enhancement, especially for document clean up and binarization and watermark removal in documents as well as in natural images. We review also some related works using the GANs in image-to-image translation. Then, we provide our proposed approach in Section 3. Some experimental results and comparison with traditional and recent methods are described in section 4. Finally,  a conclusion with some future research directions are presented in Section 5.
%I wish you the best of success.

 

\section{Related works}
In this work, we focus on the enhancement of degraded document images by addressing different kinds of degradation which are  document clean up, binarization, and watermark removal. From a document analyst viewpoint, this recovery of a clean version from the degraded document  falls in the research field called document enhancement.  Where we can find, in addition to those three, many other ways to enhance a document. For instance:  unshadowing \cite{art30,art35}, super-resolution \cite{art36},  deblurring \cite{art37}, dewarping \cite{art45}, etc. In what follows, we cover some related works to our addressed problems.  

\subsection{Degraded document recovering and binarization}

Dirty document cleaning is related to document image binarization, especially the degraded ones. Where the goal is to produce a binary but clean document. That is why, dealing with these two problems was almost the same. The idea is to classify the pixels of the document as one of two categories: degradation or text. Afterward,  assigning zeros to the text pixels and ones for the degradation will generate  a binary clean image. While generating a gray-scale or colored image is done by preserving the same value for the text pixels.

Classic document binarization methods \cite{art3,art42,art43,art4,art5,art47}  were based on thresholding, many algorithms were developed for the goal of  finding the suitable global or local threshold(s) to apply as a filter. According to the threshold(s),  pixels are classified to be  belonging on the text (zero) or the degradation (one). Lelore et al. \cite{art49} presented an algorithm called FAIR, based on edge detection to localize the text  in a degraded document image. A global threshold selection method was proposed in \cite{art51}, basing on fuzzy expert systems (FESs), the image contrast is enhanced. Then, the range of the threshold value is adjusted by  another FES and  a pixel-counting algorithm. Finally, the threshold value is obtained as the middle value of that range. A machine learning based approach was proposed in \cite{art9}, the goal was  the determination of the binarization threshold in each image region given a three-dimensional feature vector formed by the distribution of the gray level pixel values. The support vector machine (SVM) was used to classify each region into one of four different threshold values. An other and similar SVM based approach was introduced in \cite{art52}. The main drawbacks of these classic methods is that the  results are highly sensitive to   document condition. With a complex image background or a  non uniform intensity, problems occurred.

Later, evolved techniques were proposed. Moghaddam et al. \cite{art48} proposed a variational model to remove the  bleed-through from the degraded double-sided document images. For the cases where the verso side of the document is not available, a modified variational model is also introduced. By transferring the model to the wavelet domain and using the hard wavelet shrinkage,  the interference patterns was removed. Other energy based methods were also introduced. In \cite{art6}, authors considered the ink as a target and tried to detect it by maximizing an energy function. This technique was applied also for  scene text binarization \cite{art56}, which is a similar task. Similarly, Xiong et al. \cite{art63} estimated the background and subtracted it from the image  by a  mathematical morphology. Then the Laplacian energy based segmentation is performed on the enhanced document image to classify the pixels. Although these sophisticated image processing techniques, document binarization results are still unsatisfactory.
For this reason, some deep learning frameworks were recently used to tackle this problem. The goal here is  not to train a model for predicting a threshold,  it is to directly separate the foreground text from the background noise given, of course,  a considerable amount of paired data (degraded images and their binarized versions). These deep learning based models lead to better results compared to the other hand-crafted methods. Several End-to-end frameworks, based on fully convolutional neural-network (encoder-decoder way), was used to binarize and enhance the document image \cite{art54,art64,art65}. Afzal et al. \cite{art50} formulated the binarization of document images as a sequence learning problem. Hence, the 2D Long Short-Term Memory (LSTM) was employed to process a 2D sequence of pixels for the classification of each pixel as text or background. In \cite{art8}, an other Fully convolutional network was trained  with a combined Pseudo F-measure and F-measure loss function for the task of document image binarization. A method that inspires from the two previous approaches, i.e.,  a recurrent neural network based algorithm using grid LSTM cells for image binarization, as well as a pseudo F-Measure based weighted loss function could be found in \cite{art53}. Vo et al. \cite{art7} proposed a  hierarchical deep supervised network (DSN) architecture to predict the text pixels. They claimed that their architecture incorporates side layers to decrease the learning time, while taking a lot of training data.

It is to note that in this paper our object is not to binarize the document images but to clean the degraded ones and preserve them in their basic grey or colored level.  But, we will test our approach in this problem for the purpose of comparison with the state-of-the art approaches.  

\subsection{Watermark removal}

Watermark removal is also related to classical document binarization
or image matting, where the goal is to decompose a single image into background and foreground knowing that this time  the text is in the background while the watermark is in the foreground. But, this problem was not proposed in document processing. In fact, the appeared works  that deal  with watermark removal was in natural images processing. In \cite{art10}, authors used an image  inpainting algorithms to remove the watermark. Before that, a statistical method was used to detect the watermark region. Dekel et al. \cite{art11}  proposed to estimate the outline sketch and alpha matte of the same watermark from a batch of different images. Two watermarks were used in this study, the goal was testing the effectiveness of a single visible watermark to protect a large set of images. Wu et al. \cite{art12} used the generative adversarial networks \cite{art13} to remove watermark from faces images used in a biometric system. Cheng et al. \cite{art14} proposed a method based on convolutional neural networks (CNN). First,  object detection algorithms were used to detect the watermark region in natural images  and then pass it to an other model to remove the watermark. In our study, we investigate for the first time the problem of watermark removal  in document images, this leads us to compare our approach with some results obtained on  natural images for the same purpose. 

\subsection{Generative adversarial networks for image-to-image transform}

As mentioned above, GANs are now achieving impressing results in  image generation and  translation. In this paragraph, we investigate  the use of this mechanism in related problems to document processing and enhancement. This shall gives intuitions to document analysis community about exploiting GANs for these tasks. In  \cite{art38}, it was demonstrated that GANs lead to improvements in semantic segmentation.  
Ledig et al. presented SRGAN  \cite{art39}, a Generative Adversarial Network for image Super-Resolution. Through it, they achieved a photo-realistic reconstructions for large upscaling factors (4$\times$). In \cite{art39}, conditional GANs were used for several image-to-image translation tasks (these tasks are related to document enhancement), given a paired data. This work was extended to  \cite{art40}, where CycleGAN, a GAN that uses impaired data,  was proposed as a solution. An other model called "pix2pix-HD" and deals with high-resolution (e.g. 2048x1024) photorealistic image-to-image translation tasks was appeared in \cite{art62}. Furthermore, an unsupervised method for image-to-image translation was proposed in \cite{art41}, where authors train two GANs, or "DualGAN" as they denoted. In their architecture, the primal GAN learns to translate images from a  domain $U$ to a 
domain $V$, while the dual GAN learns to invert the task.
The closed loop architecture allows images from each domain to be translated and then reconstructed. Hence, a loss function that accounts for the reconstruction error of images can be used to train the translators. 

\section{Proposed approach}

\begin{figure*}[ht]
    \includegraphics[width=\textwidth]{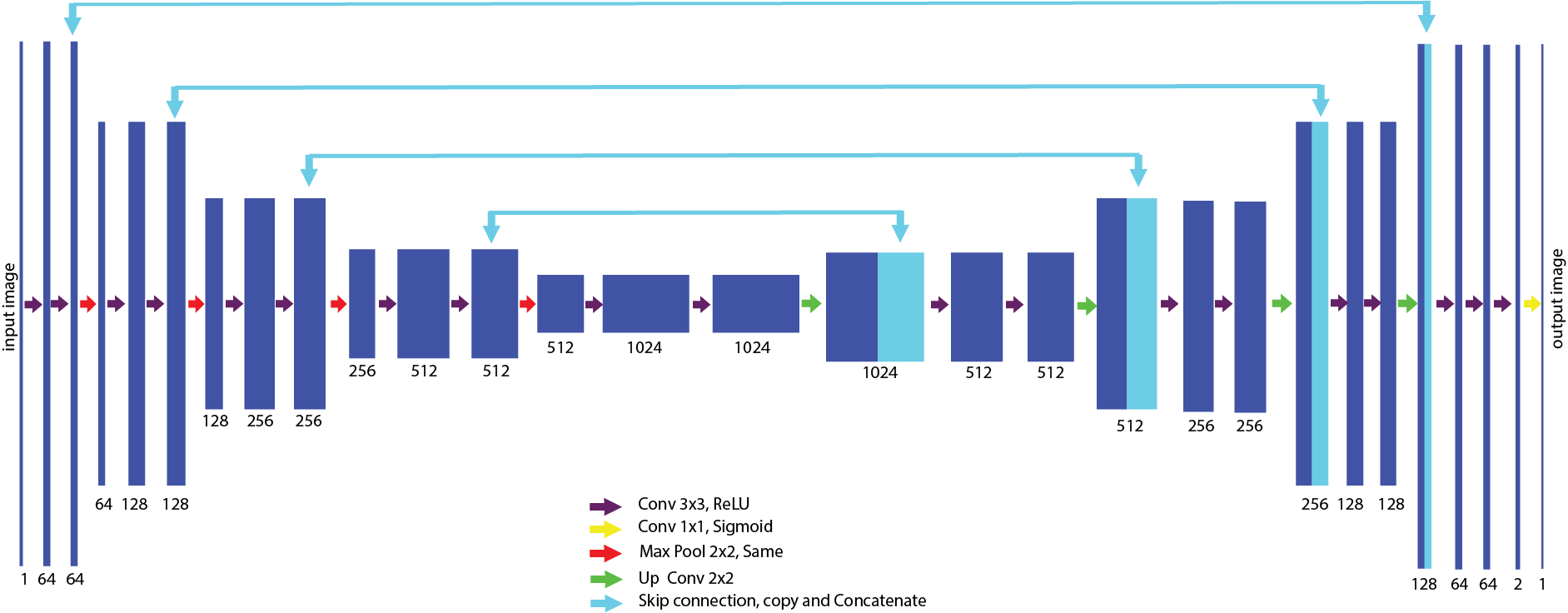}
    \caption{The generator follows the U-net architecture \cite{art17}. Each  box corresponds to a feature map. The number of channels is denoted
    on bottom of the box. The arrows denote the different operations.} \label{fig2}
    \end{figure*}

We consider the problems of document enhancement as an image to image translation task where the goal is to $generate$  clean document images given the degraded ones. Since GANs have outperformed auto-encoders in generating high fidelity 
samples and  while we are using paired data, we propose to use a cGAN. We called our model $DE$-$GAN$ (for Document Enhancement conditional Generative Adversarial Network). GANs were initially proposed in \cite{art41} and consist in two neural networks, a generator $G$ and a discriminator $D$ characterized by the parameters $\varphi_G$ and $\varphi_D$, respectively. The generator have the goal of  learning a   mapping from a random noise vector $z$ to an image $y$, $G_{\varphi_G}$: $z \rightarrow  y$. While the discriminator has the function of distinguishing between the image generated by $G$ and the ground truth one. Hence,   given $y$, $D$ should be able to tell if it is $fake$ or $real$ by outputting a probability value, $D_{\varphi_D}$: $y \rightarrow   P(real)$.  Those two networks compete against each other in a min-max game, in other words if one wins the other loses. The generator aims to cheat the discriminator by producing a close image to the ground truth, however,  the discriminator will improve his prediction of the image being fake, and this is what is called the adversarial learning. cGANs follow the same process, except that, they introduced an additional parameter $x$. Which is the conditioned image. Here, the generator is learning the mapping from an observed image $x$ and a random noise vector $z$, to $y$, $G_{\varphi_G}$: $\{x, z \} \rightarrow  y$ and the discriminator is looking, also, to the conditioned image which makes his process as: $D_{\varphi_D}$: $ \{x, y \} \rightarrow   P(real)$.

In  our situation, the generator will generate a clean image denoted by $I^C$ given the degraded (or watermarked) one which we will denote $I^W$. The generator aims, of course, to produce an image that is very close to the ground truth image denoted by $I^{GT}$. The training of cGANs for this task is done   by the following adversarial loss function:
%\begingroup\makeatletter\def\f@size{9}\check@mathfonts
\begin{dmath}
    L_{GAN} (\varphi_G,\varphi_D)= \mathbb{E}_{I^{W}, I^{GT}}  \textnormal{ log} [D_{\varphi_D}(I^{W}, I^{GT})] + \mathbb{E}_{I^{W}} \textnormal{ log} [1 - D_{\varphi_D}(I^{W}, G_{\varphi_G} (I^{W}))] 
\end{dmath}
%\endgroup

Using this function, the generator should produce, after several epochs, a similar image to the ground truth, i.e., the watermark and the degradation will be removed and this may fool the discriminator. But, it is not guaranteed that the text will be preserved in a good condition. To overcome this, we employ an additional log loss function between the generated image and the ground truth, for the purpose of forcing the model to generate images that have the same text as the ground truth. It is to note also that this additional loss boosts the training speed, the added function is: 

\begin{dmath}
    L_{log} (\varphi_G) = \mathbb{E}_{I^{GT}, I^{W}} [-( I^{GT} \textnormal{ log}(G_{\varphi_G} (I^{W}))+((1-I^{GT})\textnormal{ log}(1-G_{\varphi_G} (I^{W})))]
\end{dmath}

Thus, the proposed loss  of our network, denoted by $L_{net}$ becomes:
% \begingroup\makeatletter\def\f@size{9}\check@mathfonts
\begin{equation}\label{eq:3}
    L_{net}(\varphi_G,\varphi_D) = min_{\varphi_G} max_{\varphi_D} L_{GAN} (\varphi_G,\varphi_D) + \lambda L_{log} (\varphi_G)  
\end{equation}
% \endgroup
Where, $\lambda$ is a hyper-parameter that was set to 100 for text cleaning and 500 in watermark removal and document binarization. The architecture  of generator and discriminator networks are described in the next sections.  

\subsection{Generator:}
    
    The generator is performing an image-to-image translation task. Usually, auto-encoder models are used  for this problem \cite{art44,art15,art16}. These models consist, mostly, in a sequence of convolutional layers called encoder which perform down-sampling until a particular layer. Then,  the process is reversed to a sequence of up-sampling and convolutional layers called decoder. There are two disadvantages of using an encoder-decoder model for the proposed problem: First, due to down-sampling (pooling), lot of information is lost and the model will have difficulties  to recover them later when predicting an image with the same size as the input. Second,  image  information flow pass through all the layers, including the bottleneck. Thus, sometimes, a huge amount of unwanted  redundant features (inputs and outputs are sharing a lot of same pixels) are exchanged. Which leads to energy and time loss. For this reason, we employ skip connections following the structure of a the model called U-net \cite{art17}.  Skip connections are added every two layers to recover images with less deterioration, it is to note also that skip connections are used when training a very deep model to prevent the gradient vanishing and exploding problems. Some batch normalization layers are also added to accelerate the training. The architecture of the generator used in this study is illustrated in Fig. \ref{fig2}, it is  similar to \cite{art17} where it was introduced for the purpose of biomedical image segmentation.

 \subsection{Discriminator}

\begin{figure}[h]
    \centering
    \includegraphics[scale=0.15]{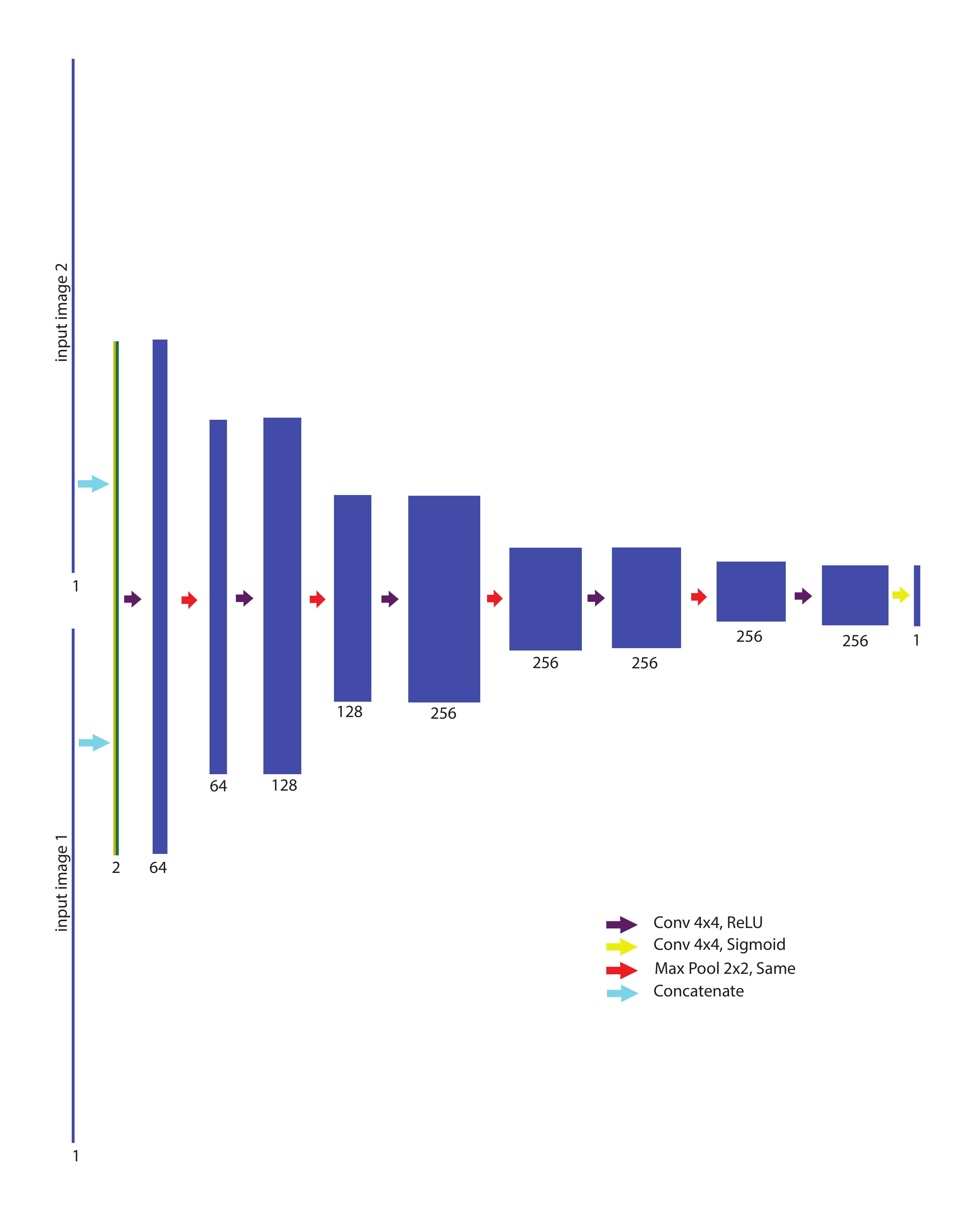}
    \caption{The Discriminator architecture} \label{fig4}
    \end{figure}

The defined discriminator is a simple  Fully Convolutional  Network (FCN),  composed of 6 convolutional layers, that output  a 2D matrix containing  probabilities of the generated image being real. This model is presented in Fig. \ref{fig4}. As shown, the discriminator receives two input images which are the degraded image and its clean version (ground truth or cleaned by the generator). Those images were concatenated together in a $256\times 256 \times 2 $ shape tensor. Then, the obtained volume propagated in the model to end up in a $16 \times 16 \times 1$ matrix in the last layer. This matrix contains probabilities that should be, to the discriminator, close to 1 if the clean image represents the ground truth. If it is generated by the generator the probabilities should be close to 0. Therefore, the last layer  takes a sigmoid  as an activation function. After completing training, this discriminator is no longer used. Given a degraded image, we only use the generative network to enhance it. But, this discriminator shall force the generator during training to produce  better results.
			
\subsection{Training process}
\begin{figure}[h]
    \centering
    \includegraphics[width=80mm, height=40mm]{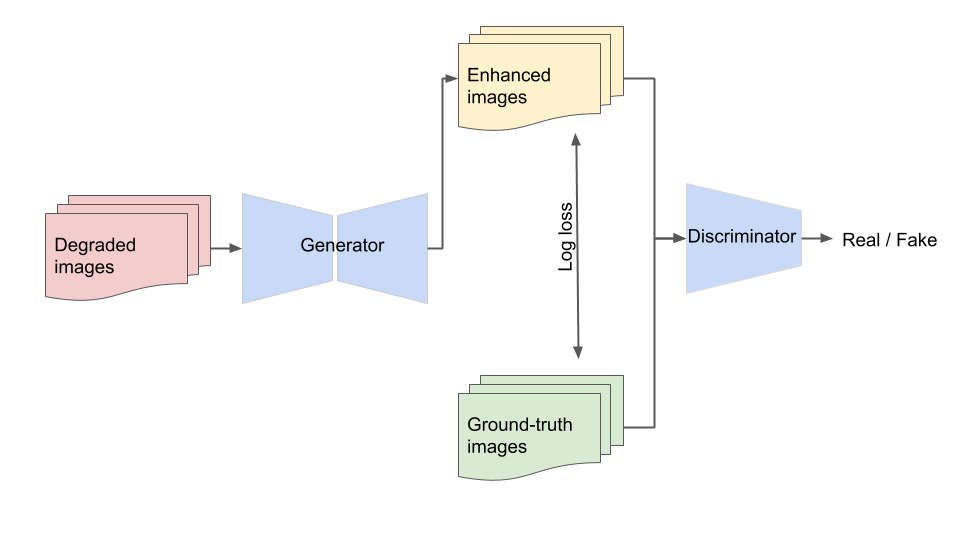}
    \caption{The proposed DE-GAN} \label{fig3}
\end{figure}

Training our DE-GAN was as follows, we took patches from the degraded images of size $256 \times 256$ and fed it as an input to the generator. The produced images are fed to the discriminator with the ground truth patches and the degraded ones. Then, as presented in  equation \ref{eq:3}, the discriminator   starts forcing the generator to produce outputs that cannot be distinguished from “real” images, while doing his best  at detecting the generator’s “fakes”. This training is illustrated in Fig. \ref{fig3} and it is done   using Adam with a learning rate of $1e^{-4}$ as an optimizer.

\section{Experiments and results}
\subsection{Document cleaning and binarization}

We begin our experiments with document cleaning. For this task, the Noisy Office Database which contains  different types of degradation, and presented in \cite{art19}, is used. We defined 112 images for training and 32 for testing. From the 112 training images, a set of overlapped patches of size $256 \times 256$ pixels was extracted. This has generated 1356 pairs of patches that were fed to our model. This first test intend to demonstrate  the adversarial training effect. Thus, we train another model which is a simple FCN which is the U-net presented in Fig. \ref{fig2}. A validation set of 15 \% from the training images was used in this model. The results obtained by both models are presented in Table \ref{tab1}.   As could be interpreted, the result of the encoder-decoder network (U-net) are acceptable for denoising and cleaning tasks. But, our DE-GAN is further improving the results. Which expose the reason of using an adversarial training for these types of problems. For more comparison, we have participated to the kaggle competition on denoising dirty documents \footnote{\href{https://www.kaggle.com/c/denoising-dirty-documents/}{https://www.kaggle.com/c/denoising-dirty-documents/}}, we obtain a root mean squared error score of 0.01952. This makes our method as one of the best approaches in the leaderboard.

\begin{table}[h]
\centering
\caption{The obtained results of document cleaning using Noisy office database \cite{art19}}\label{tab1}
\begin{tabular}{|l|l|l|}
\hline
Model &  SSIM & PSNR\\
\hline
FCN (U-net) &  0.9970 & 36.02\\
\hline
\textbf{DE-GAN}  & \textbf{0.9986} & \textbf{38.12}\\
\hline
\end{tabular}
\end{table}

\begin{figure}[h]
\centering
\includegraphics[scale=0.12]{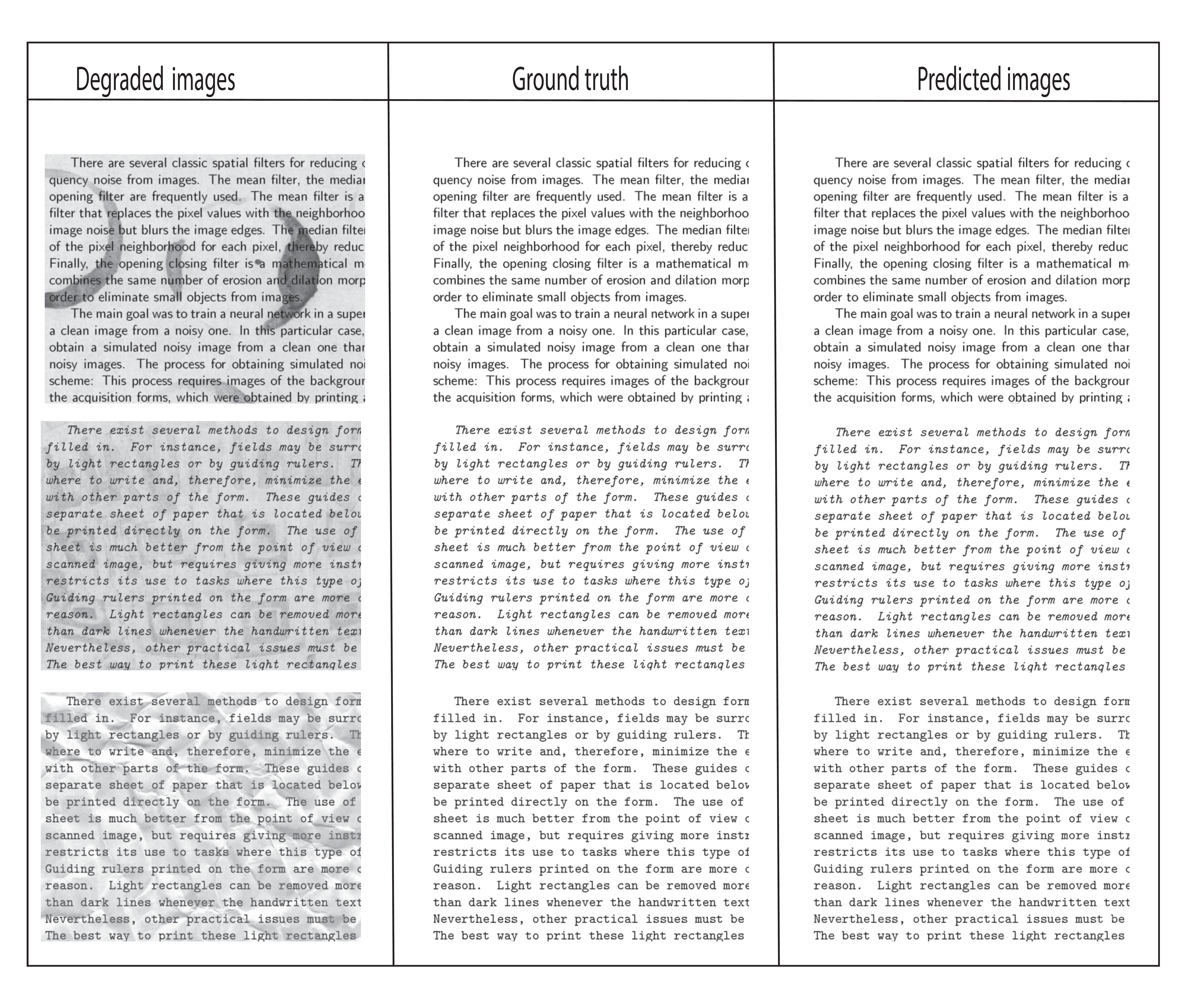}
\caption{Cleaning degraded documents by DE-GAN} \label{fig5}
\end{figure}

In order to give an idea of the cleaning made by our model, some examples are given in Fig. \ref{fig5} which demonstrate the ability of recovering a  very close document to the ground truth.

 Nevertheless, we will compare our approach with state-of-the-art results in the document binarization problem. We take the  DIBCO 2013 Dataset \cite{art20} for testing. While training our model was done with different versions of  DIBCO Databases \cite{art22,art23,art24,art21,art25,art26}. Same as the previous test, a set of 6824 training pairs (patches of size $256 \times 256$) was taken from its 80 total images. The obtained results are compared with several approaches  in Table \ref{tab2}.

\begin{table}[h]
\centering
\caption{Results of image binarization on DIBCO 2013 Database.}\label{tab2}
\begin{tabular}{|l|l|l|l|l|}
\hline
Model  & PSNR & F-measure & F$_{ps}$&DRD\\
\hline
Otsu\cite{art3} & 16.6 & 83.9 &86.5& 11.0\\
\hline
Niblack\cite{art43} & 13.6 & 72.8 &72.2& 13.6\\
\hline
Sauvola et al.\cite{art42} & 16.9 & 85.0&89.8& 7.6\\
\hline
Gatos et al. \cite{art29} & 17.1 & 83.4&87.0&9.5\\
\hline
Su et al. \cite{art28} & 19.6 & 87.7&88.3&4.2\\
\hline
Tensmeyer et al \cite{art8} & 20.7 &93.1& 96.8 &2.2\\
\hline
Xiong et al. \cite{art63} & 21.3 & 93.5&94.4&2.7\\
\hline
Vo et al. \cite{art7} & 21.4 & 94.4&96.0&1.8\\
\hline
Howe \cite{art27} & 21.3 & 91.3&91.7&3.2\\
\hline
\textbf{DE-GAN}  &   \textbf{24.9} & \textbf{99.5}&\textbf{99.7}& \textbf{1.1}\\
\hline
\end{tabular}
\end{table}
    
\begin{figure}[h]
\centering
\includegraphics[scale=0.091]{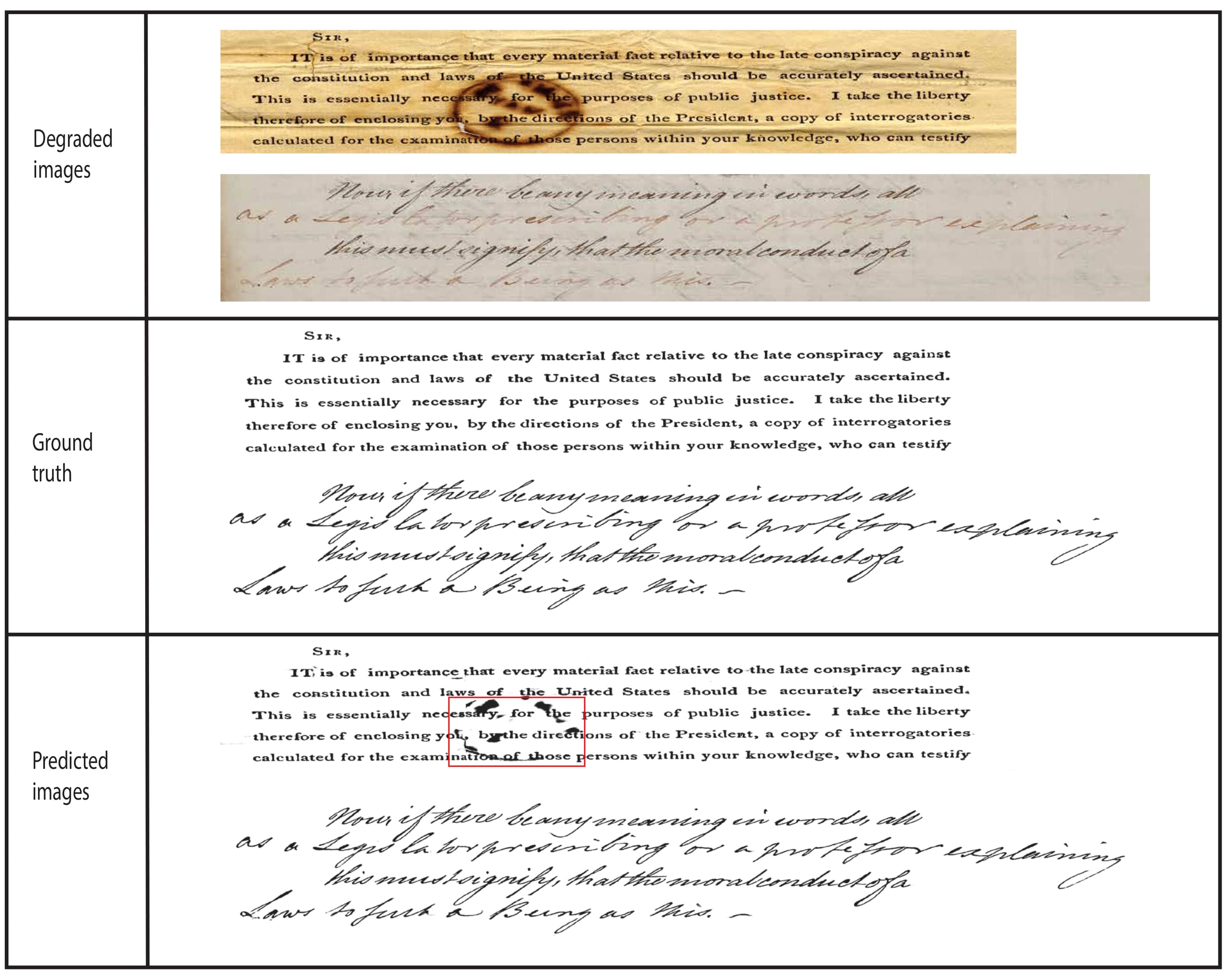}
\caption{Binarization of degraded documents by DE-GAN, the result is satisfactory, except in some parts that were highly dense (the red boxes in the predicted images row)} \label{fig6}
\end{figure}

 Out of the results, we can say that  DE-GAN is superior than the current state-of-the-art methods according to the following metrics   \cite{art20}: Peak signal-to-noise ratio (PSNR), F-measure, pseudo-F-measure (F$_{ps}$) and Distance reciprocal distortion metric (DRD). Some examples of DIBCO 2013 images binarization by DE-GAN are presented in Fig. \ref{fig6}. 

\begin{figure}[h]
\centering
\begin{tabular}{cc}
  \includegraphics[width=40mm, height=20mm]{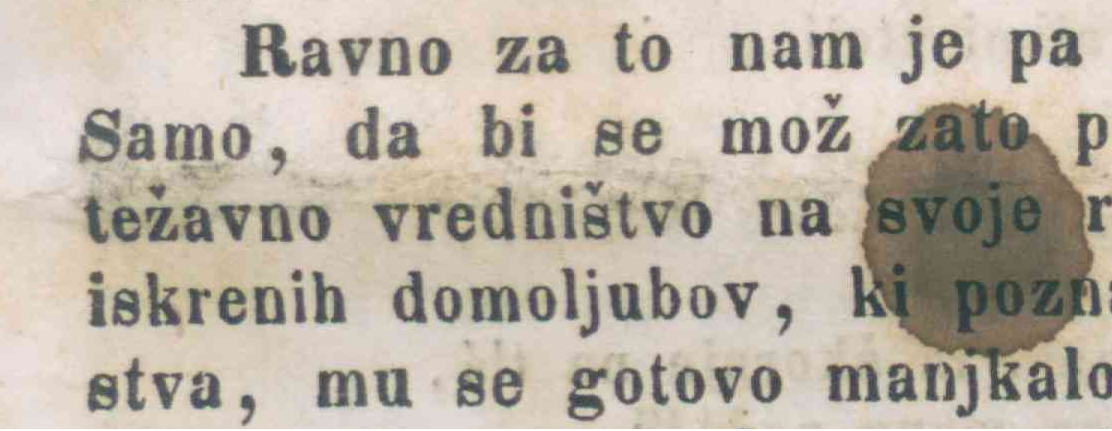} &   \includegraphics[width=40mm, height=20mm]{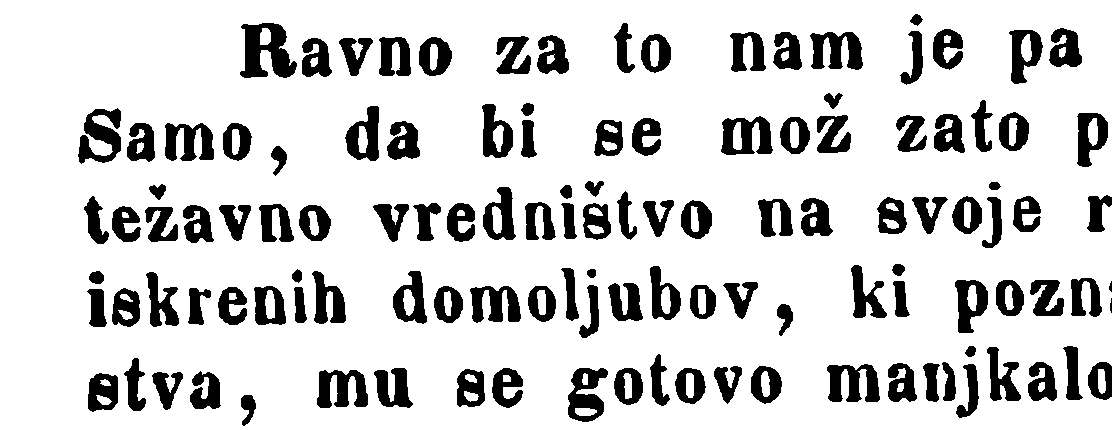}   \\  
  Original & Ground truth \\[6pt]
  \includegraphics[width=40mm, height=20mm]{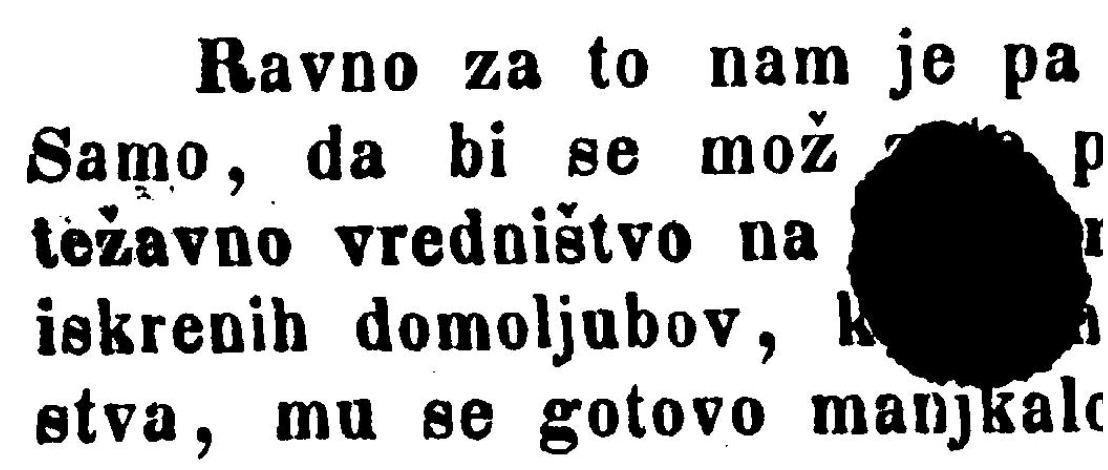} &
  \includegraphics[width=40mm, height=20mm]{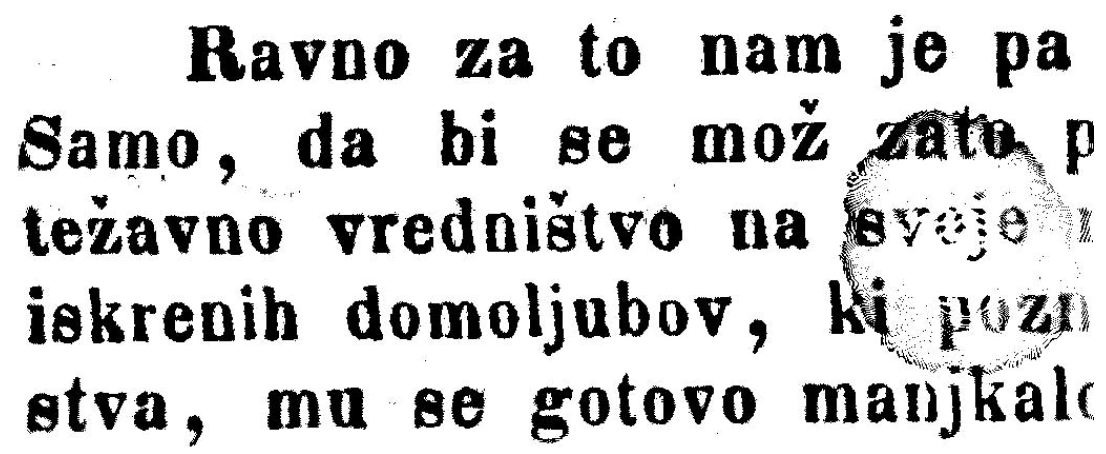}\\
  Otsu\cite{art3}  & Niblack\cite{art43} \\[6pt]
 \includegraphics[width=40mm, height=20mm]{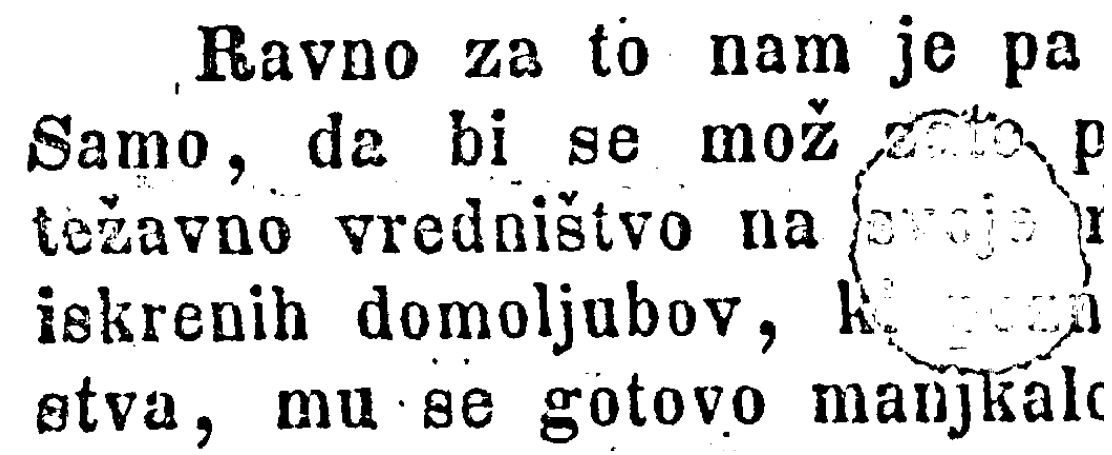}  &   \includegraphics[width=40mm, height=20mm]{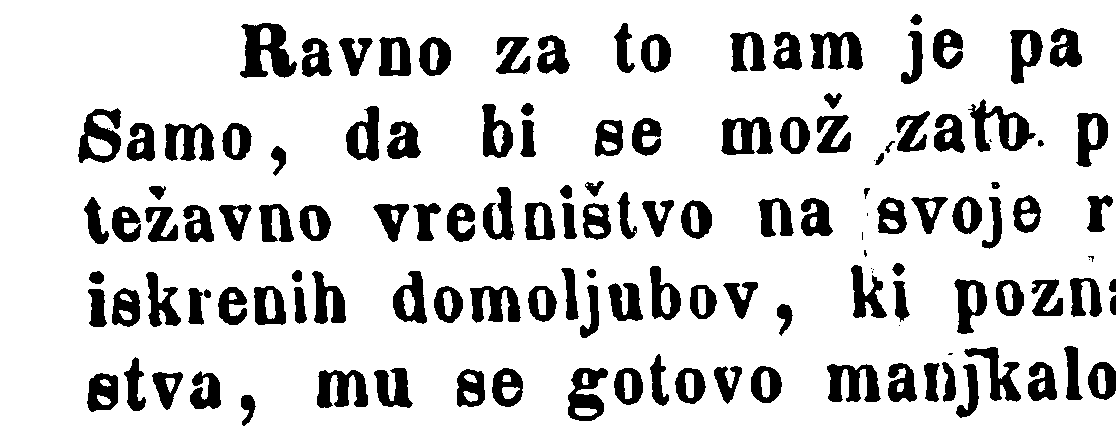} \\
  Sauvola et al.\cite{art42} & DE-GAN  \\[6pt]
\end{tabular}
\caption{Qualitative binarization results  produced by different methods of a part from the sample (PR5), which is included in DIBCO 2013 dataset}
\label{figcomp1}
\end{figure}

\begin{figure}[h]
\centering
\begin{tabular}{cc}
  \includegraphics[width=40mm, height=22mm]{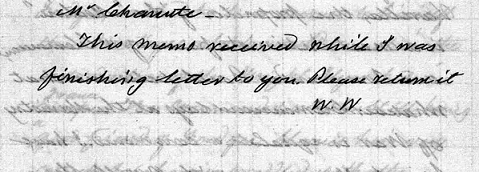} &   \includegraphics[width=40mm, height=22mm]{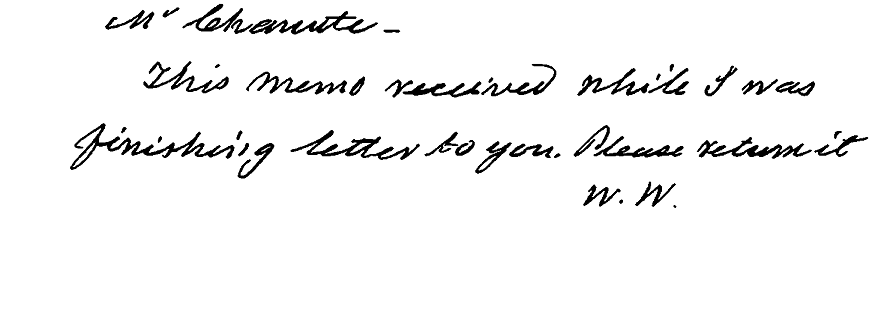}   \\
  Original & Ground truth\\[6pt]
  \includegraphics[width=40mm, height=22mm]{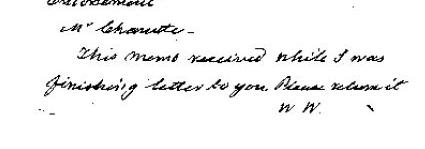} & \includegraphics[width=40mm, height=22mm]{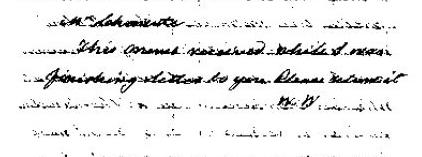} \\
  Otsu\cite{art3} & Howe \cite{art27}\\[6pt]
\includegraphics[width=40mm, height=22mm]{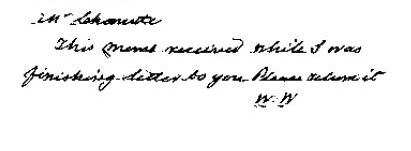}  &   \includegraphics[width=40mm, height=22mm]{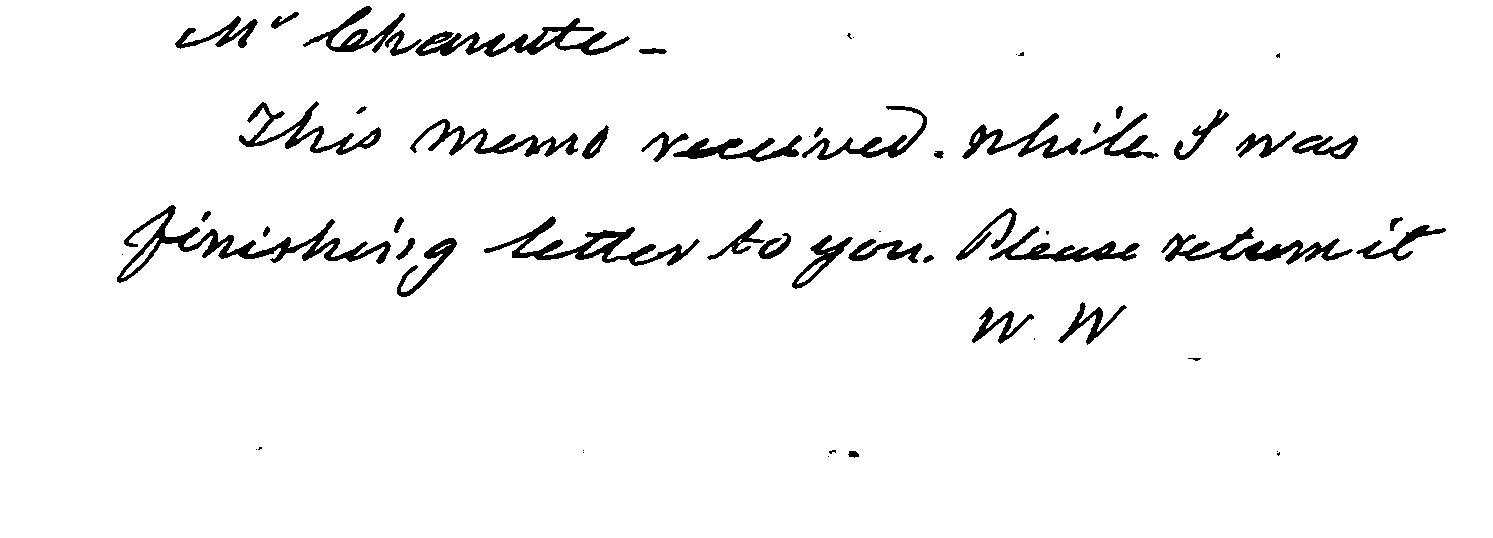} \\
 Vo et al. \cite{art7} & DE-GAN  \\[6pt]
\end{tabular}
\caption{Qualitative binarization results  produced by different methods of of a part from the sample (HW5), which is included in DIBCO 2013 dataset}
\label{figcomp2}
\end{figure}

To reflect the results of the previous Table, an illustrative comparisons between those different methods could be found in Fig. \ref{figcomp1} and Fig. \ref{figcomp2}. It is easy to visualize the superiority of our method  over the classic methods, like those of \cite{art3,art43,art42}, which fail to remove the background degradation from the document when it get very dense, because they are basing on  thresholds that make the degraded pixels classified as a text, or classifying the text pixels as a damage to be removed. For the recent approaches \cite{art27, art7}, they yield a better result than the classic ones and separate the text from the background successfully. However, our method gives a  higher performance in terms of closeness to the ground truth image.    
 
%%%%%%%%%%%%%%%%%%%%%%%%%%%%%%%%%%%%%

\begin{figure}[h]
\centering
\begin{tabular}{cc}
  \includegraphics[width=43mm, height=20mm]{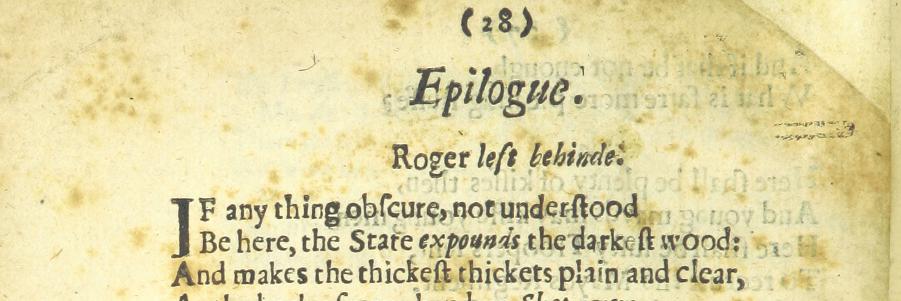}&
  \includegraphics[width=43mm, height=20mm]{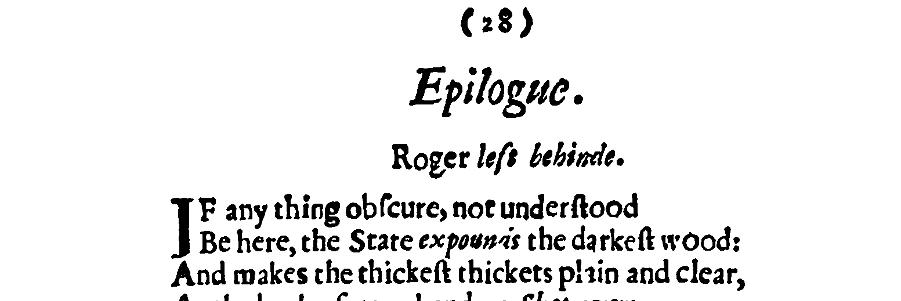}\\
Original &Ground truth\\[6pt]
  \includegraphics[width=43mm, height=20mm]{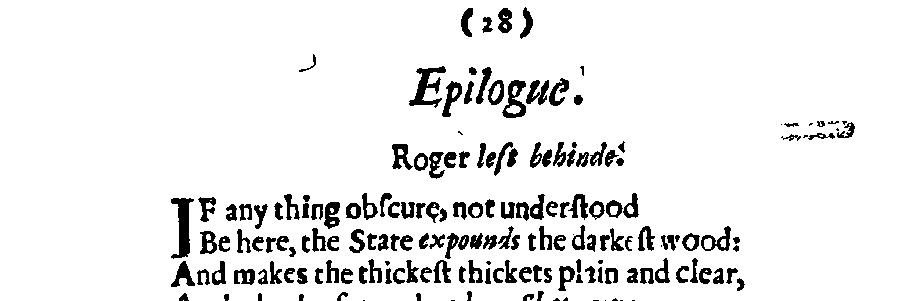}&
  \includegraphics[width=43mm, height=20mm]{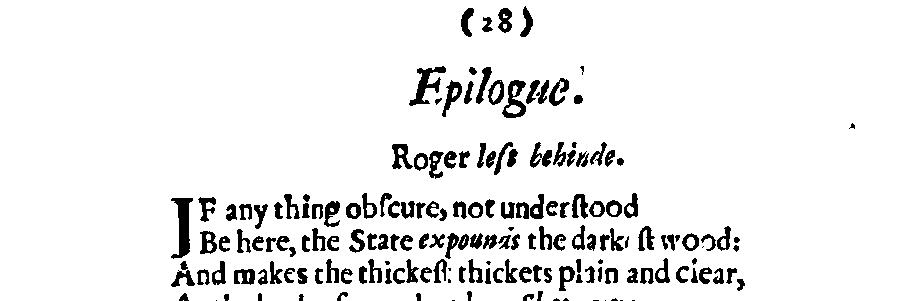}\\
Winner's output & DE-GAN \\[6pt]
\end{tabular}
\caption{Qualitative binarization results  produced by different of the sample 16 in DIBCO 2017 dataset, here we compare DE-GAN with the winner's approach.}
\label{figcomp2017}
\end{figure}

Moreover, we tested DE-GAN on a recent DIBCO dataset, which is DIBCO 2017 \cite{art26}.  We train our model on 6098 patches from similar datasets \cite{art20,art22,art23,art24,art21,art25}. The comparison is done with the  top  5 ranked  approaches in ICDAR 2017 competition on document image Binarization \cite{art26}. 18 research groups have participated in the competition with 26 distinct algorithms.  The results are  presented in table \ref{tab2017}, where you can notice the superiority of our DE-GAN over the different methods. It is to note that most of these approaches are based on encoder-decoder models and the winner team was using a U-net with several data augmentation techniques. However, GANs were not exploited in this competition. An example to compare our output with the winner algorithm is given in figure \ref{figcomp2017}.

\begin{table}[h]
\centering
\caption{Results of image binarization on DIBCO 2017 Database, a comparison with DIBCO 2017 competitors approaches.}\label{tab2017}
\begin{tabular}{|l|l|l|l|l|l|}
\hline
Model  & PSNR & F-measure & F$_{ps}$&DRD&\vtop{\hbox{\strut Rank in the}\hbox{\strut competition}}\\
\hline
10 \cite{art26} & 18.28 & 91.04 &92.86& 3.40&1\\
\hline
17a \cite{art26} & 17.58 & 89.67 &91.03& 4.35&2\\
\hline
12 \cite{art26} & 17.61 & 89.42 &91.52& 3.56&3\\
\hline
1b \cite{art26} & 17.53 & 86.05 &90.25& 4.52&4\\
\hline
1a \cite{art26} & 17.07& 83.76 &90.35& 4.33&5\\
\hline
\textbf{DE-GAN}  &   \textbf{18.74} & \textbf{97.91}&\textbf{98.23}& \textbf{3.01}&-\\
\hline
\end{tabular}
\end{table}

\begin{figure}[h]
\centering
\begin{tabular}{c}
  \includegraphics[width=80mm, height=25mm]{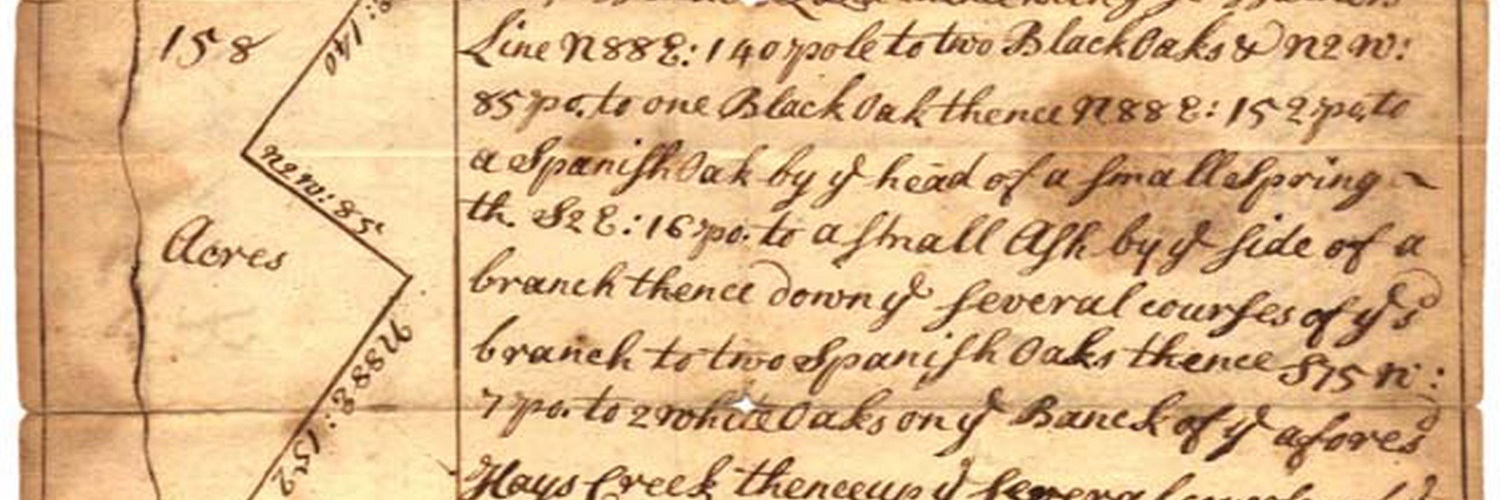}\\
  \includegraphics[width=80mm, height=25mm]{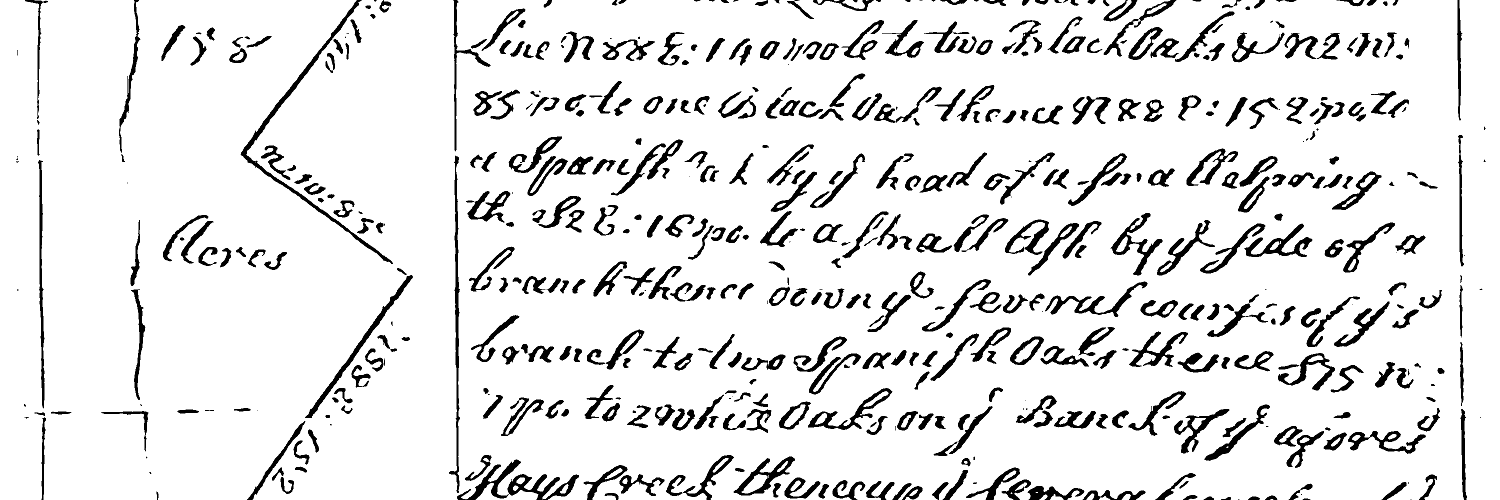}\\[0.5 cm]
  
    \includegraphics[width=80mm, height=25mm]{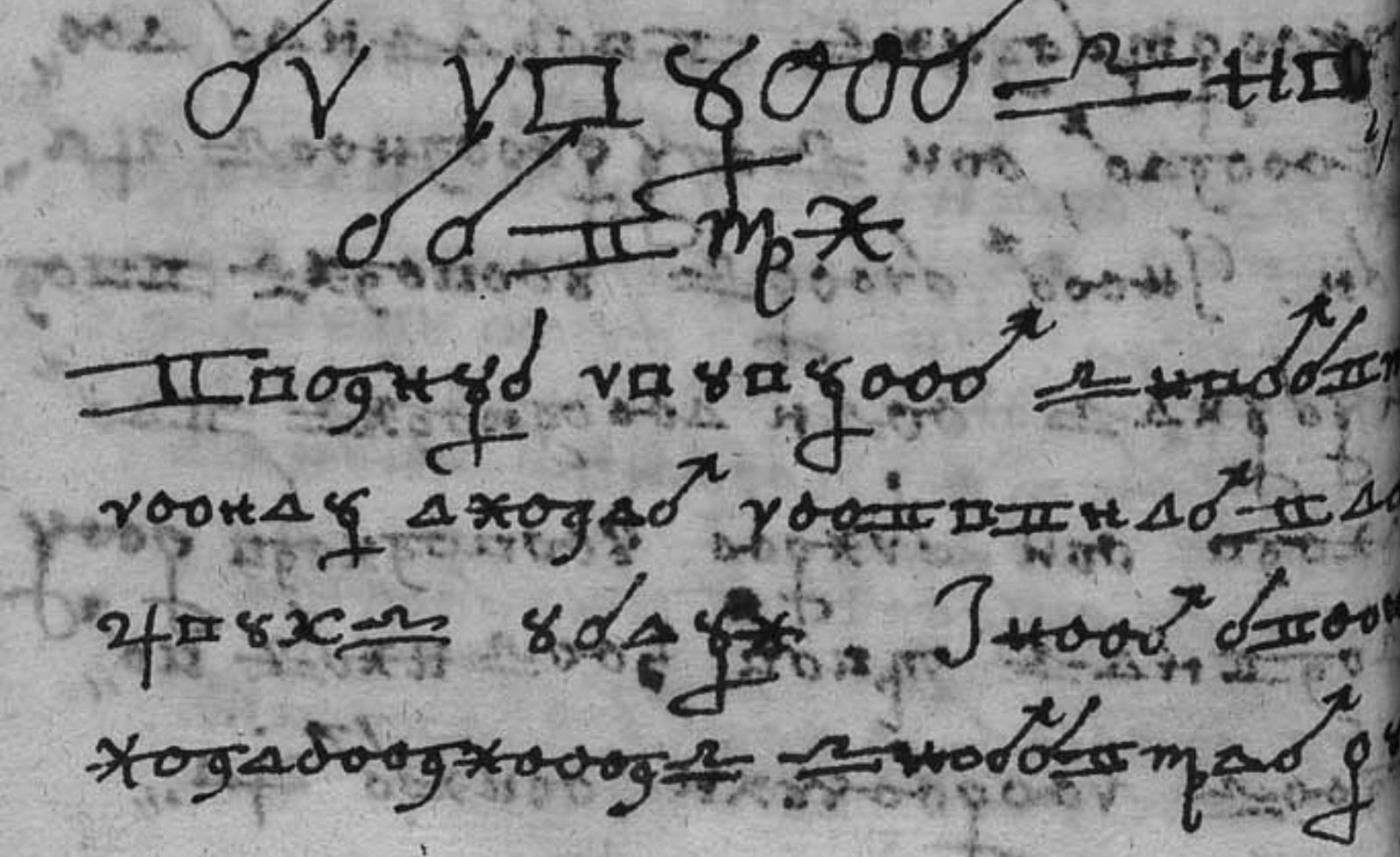}\\
  \includegraphics[width=80mm, height=25mm]{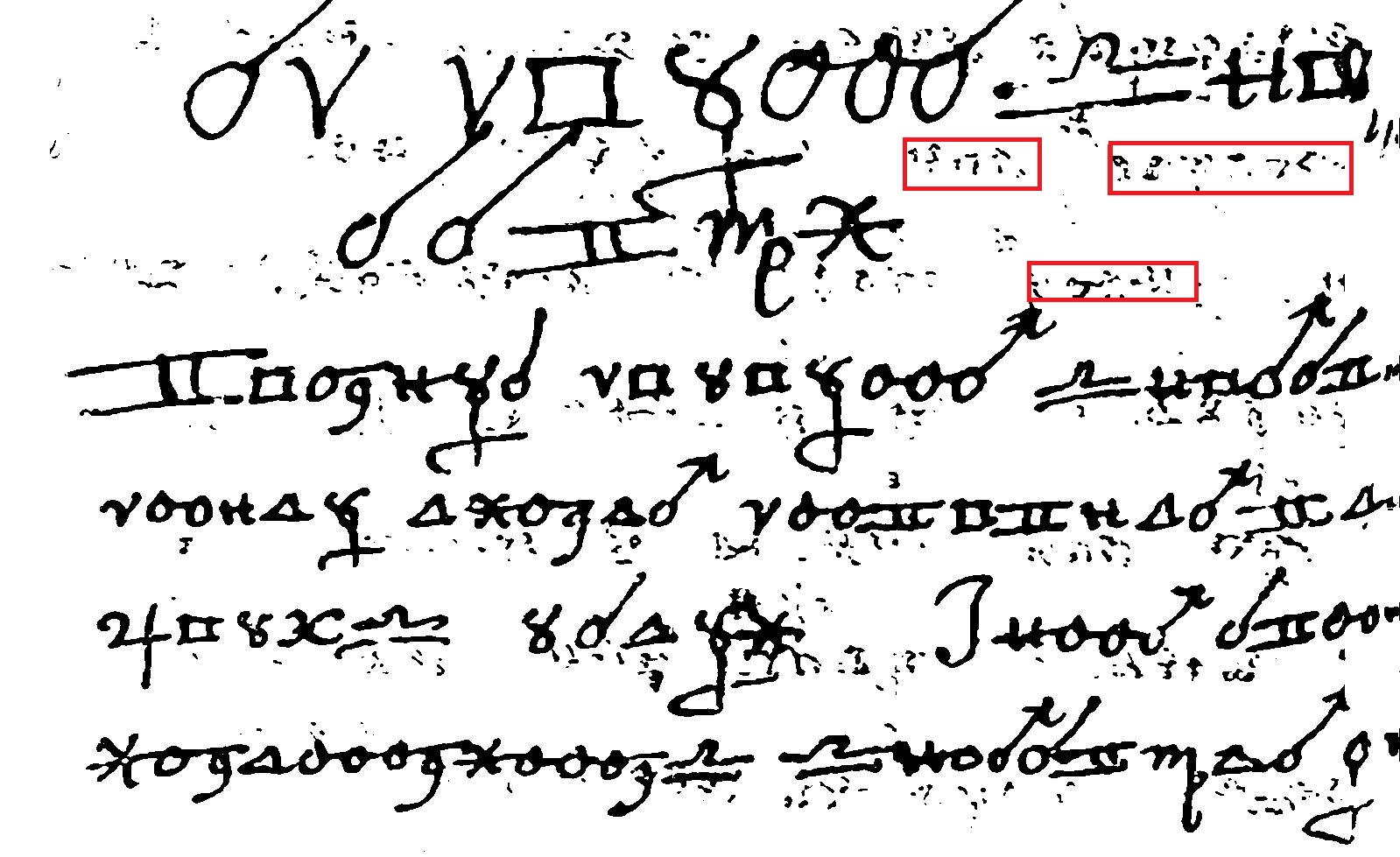}\\[0.5 cm]
  
    \includegraphics[width=80mm, height=25mm]{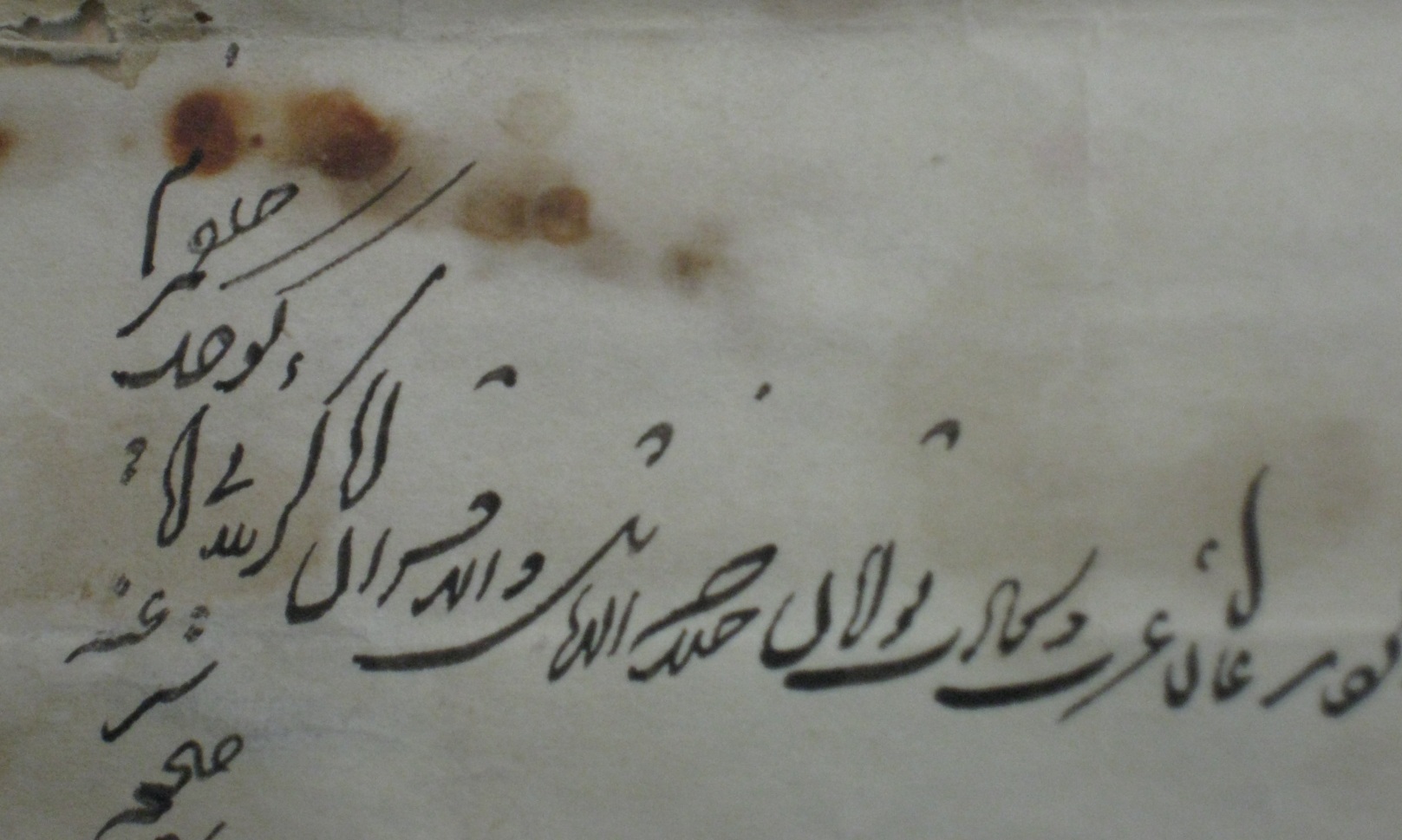}\\
  \includegraphics[width=80mm, height=25mm]{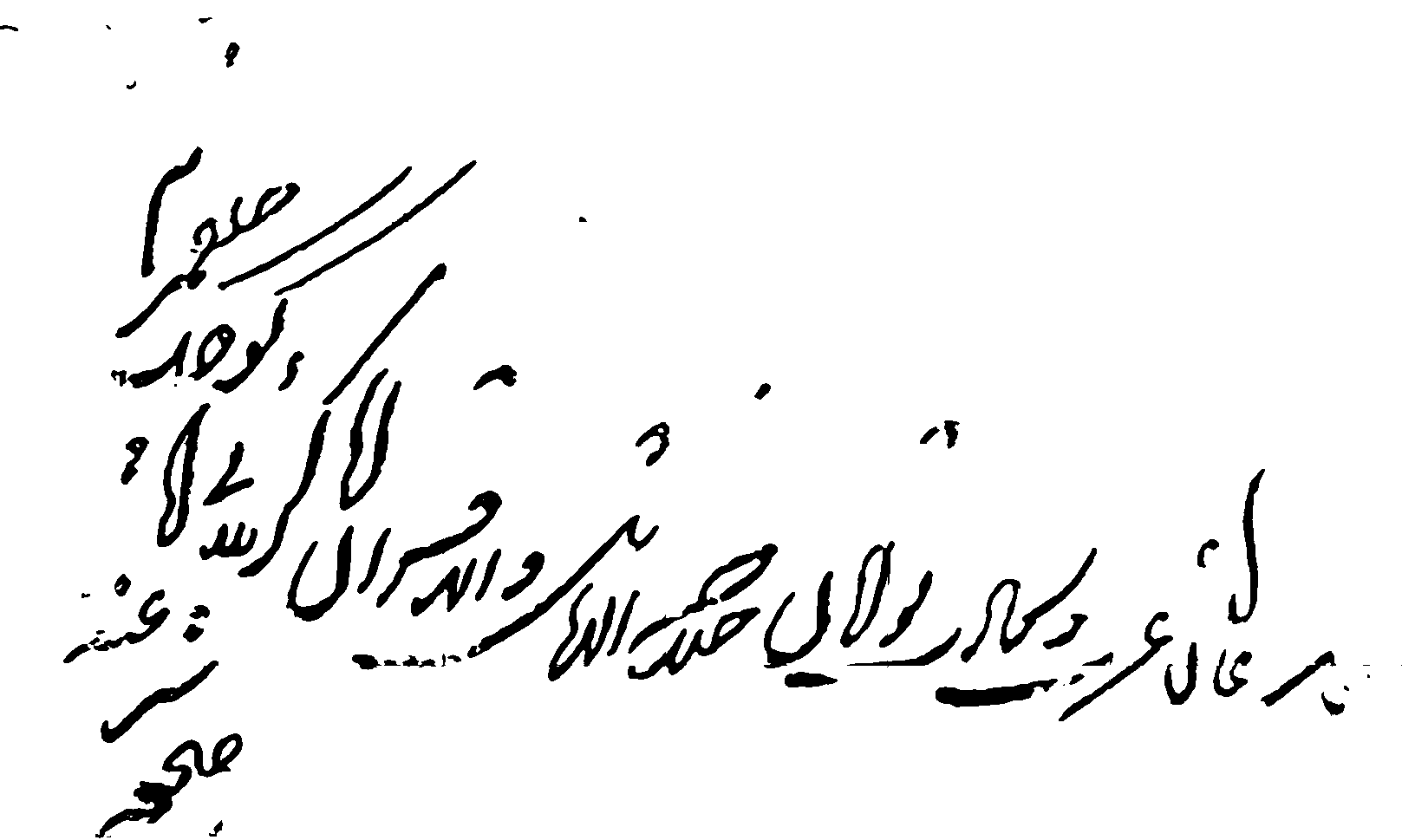}\\[0.5 cm]

\end{tabular}
\caption{Binarization of three historical degraded documents by DE-GAN, the binarized version is presented under each original image. Some parts are not well recovered as shown in the red boxes. }
\label{figreal}
\end{figure}

 In addition, we compared our model with the most recent approaches, presented in the  H-DIBCO 2018 competition \cite{art58} that was held in ICFHR 2018 conference. The results are presented in Table \ref{tab2018}. As shown, our approach has the best performance on DIBCO 2017 test set and gives the second best DRD, PSNR, F-measure and pseudo F-Measure on H-DIBCO 2018 test set. We note that the winner system in the competition integrates a lot of pre-processing and a post-processing steps in their algorithm, that make it more efficient for this particular H-DIBCO 2018 dataset. On the contrary, we are presenting a simple end-to-end model that shows a good ability in a several datasets and enhancements tasks without any additional processing step. Finally,  for a more practical  usage of the model,  we tried  to binarize some real (naturally degraded) documents as well,  the degradation consists in stains and show-through . The obtained  results are given in Fig. \ref{figreal}, the model is producing a better versions of the real images, which will certainly improves their recognition rate.

\begin{table*}[h]
\centering
\caption{Results of image binarization on DIBCO 2017 and DIBCO 2018 Databases, a comparison with DIBCO 2018 competitors approaches.}\label{tab2018}
\begin{tabular}{|l|l|l|l|l|l|l|l|l|l|}
\hline
\multirow{2}{*}{Model}  &  \multicolumn{4}{|l|}{DIBCO 2018} & \multicolumn{4}{|l|}{DIBCO 2017}& \multirow{2}{*}{\vtop{\hbox{\strut Rank in the}\hbox{\strut competition}}}\\
\cline{2-9}
{}& PSNR & F-measure & F$_{ps}$&DRD&PSNR & F-measure & F$_{ps}$&DRD&{}\\
\hline
1 \cite{art58} & \textbf{19.11} & \textbf{88.34} &\textbf{90.24}& \textbf{4.92}& 17.99 & 89.37 &90.17& 5.51&1\\
\hline
7 \cite{art58} & 14.62 & 73.45 &75.94& 26.24& 15.72 & 84.36 &87.34& 7.56&2\\
\hline
2 \cite{art58} & 13.58 & 70.04 &74.68& 17.45& 14.04 & 79.41 &82.62& 10.70&3\\
\hline
3b \cite{art58} & 13.57 & 64.52 &68.29& 16.67& 15.28 & 82.43 &86.74& 6.97&4\\
\hline
6 \cite{art58} & 11.79 & 46.35 &51.39& 24.56& 15.38 & 80.75 &87.24& 6.22&5\\
\hline
DE-GAN  & 16.16 & 77.59 &85.74& {7.93}&  \textbf{18.74} & \textbf{97.91}&\textbf{98.23}& \textbf{3.01}&-\\
\hline
\end{tabular}
\end{table*}

\subsection{Watermark removal}
After testing our model in document cleaning and binarization, we will evaluate it on the problem of  watermark removal. Dense watermarks (or stamps) can cause a huge deterioration in the $foreground$ of the document, which make it hard to be read. However, this problem was not investigated by document analysis community.  We decided to be the first that address  it using DE-GAN.  Hence, it was not possible to find a public dataset for testing.  We created our own database which contains $1000$ pairs (image of a document with a dense watermark and stamps and its clean version). 
%\begin{figure}[ht]
%\centering
%\includegraphics[width=70mm, height=60mm]{images/stamps.jpg}
%\caption{The degradation caused by dense stamps} \label{figstamps}
%\end{figure}   

\begin{figure}[ht]
\centering
\begin{tabular}{cc}
  \includegraphics[width=35mm, height=50mm]{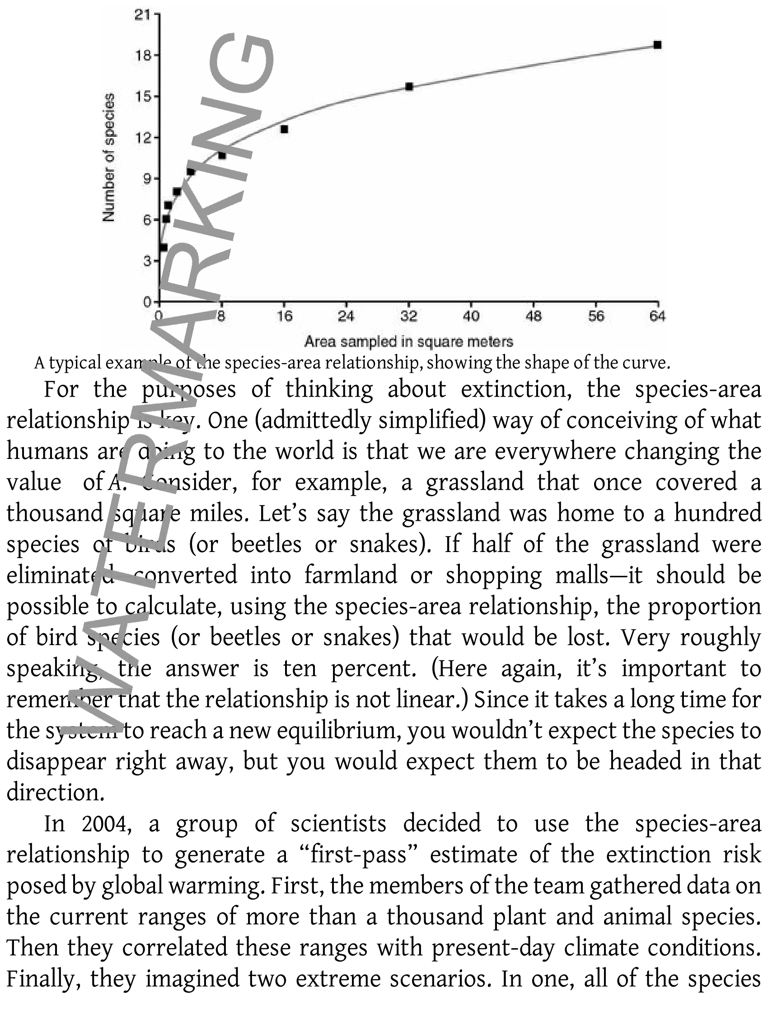} &   \includegraphics[width=35mm, height=50mm]{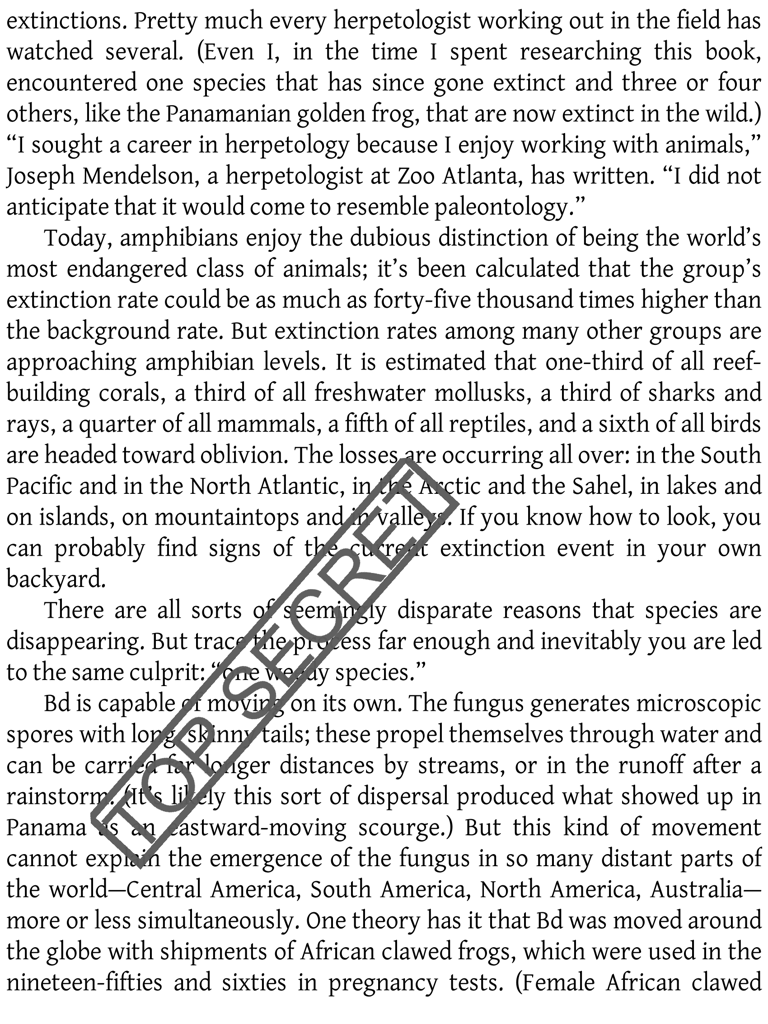}  \\
\includegraphics[width=35mm, height=50mm]{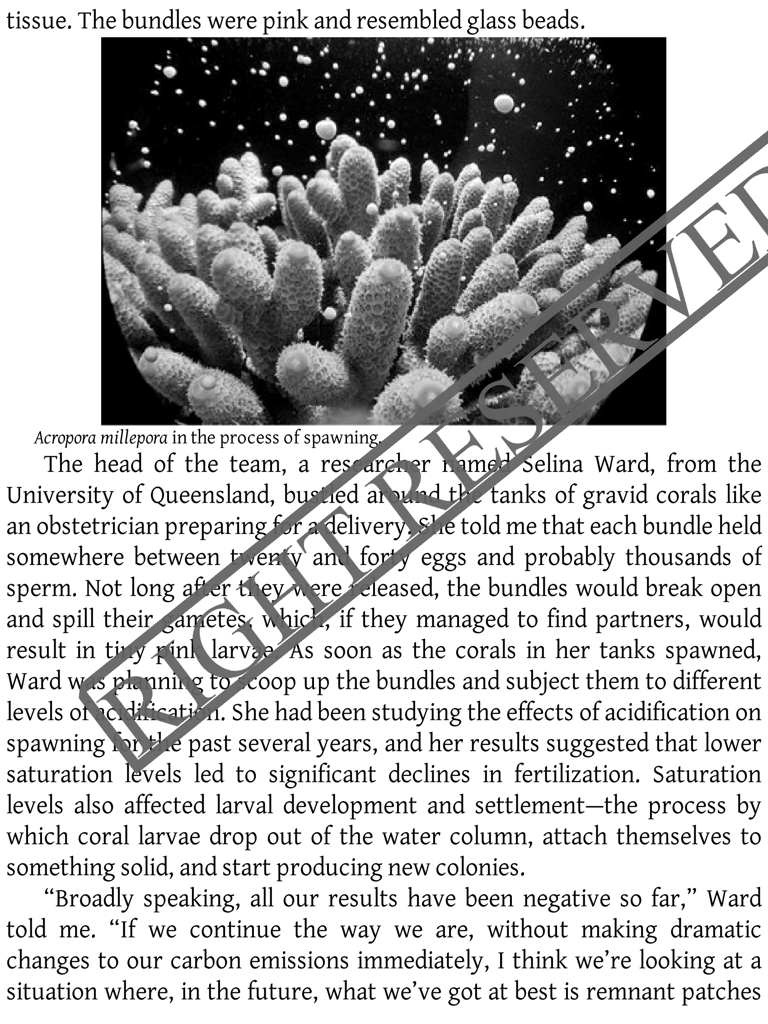} &
\includegraphics[width=35mm, height=50mm]{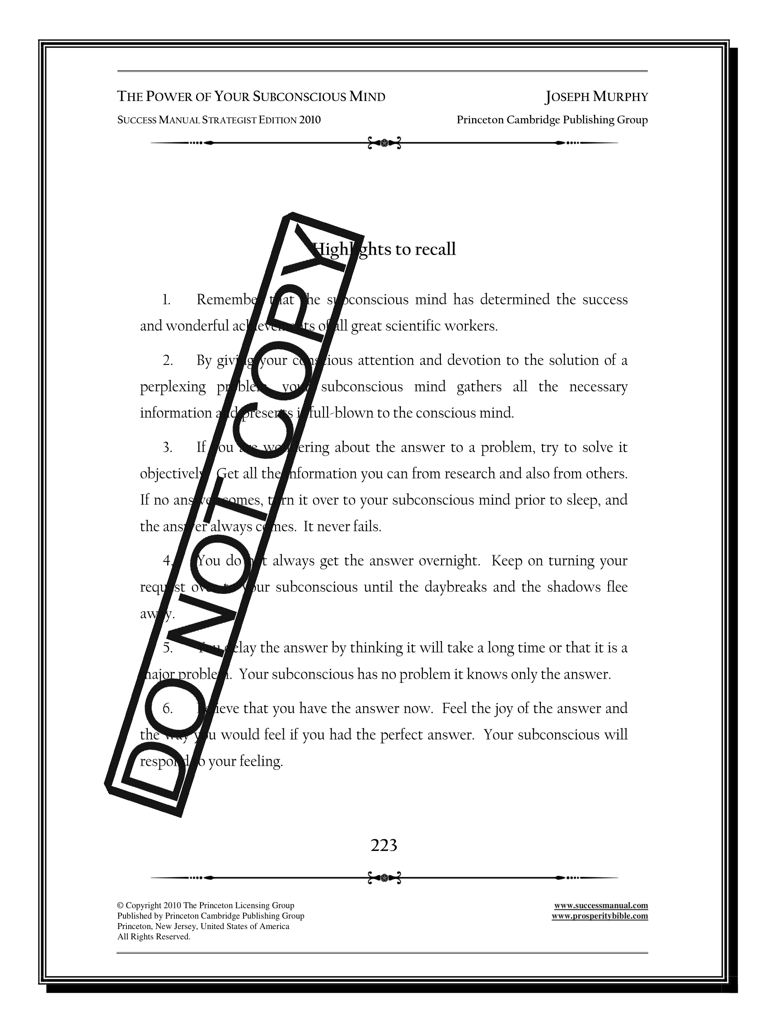}\\
\end{tabular}
\caption{4 Samples from our developed Dataset}
\label{figsamples}
\end{figure}

The used watermarks have random texts, sizes, colors, fonts, opacities and locations (see Fig. \ref{figsamples}). As shown, these watermarks are sometimes covering the entire text making it unseen by the unaided eye.  The code used to produce this data is available at GitHub \footnote{\href{https://github.com/dali92002/watermarking-documents/blob/master/Watermarking.ipynb}{https://github.com/dali92002/watermarking-documents/blob/master/Watermarking.ipynb}}, for the same dataset used in our study the reader can contact the first author to obtain it.  Training our DE-GAN was done, same as document cleaning, by using  overlapped patches ($7658$ pairs of patches from $800$ watermarked document images). While taking $200$ documents for testing.  Since, for the best of our knowledge, there is no approach in the literature that addresses this problem in documents. Comparing our obtained results  was  done with the approaches used in natural images watermark removal. The comparison results are presented in Table \ref{tab3}.

\begin{figure}[h]
\centering
\begin{tabular}{c}

    \includegraphics[width=80mm, height=50mm]{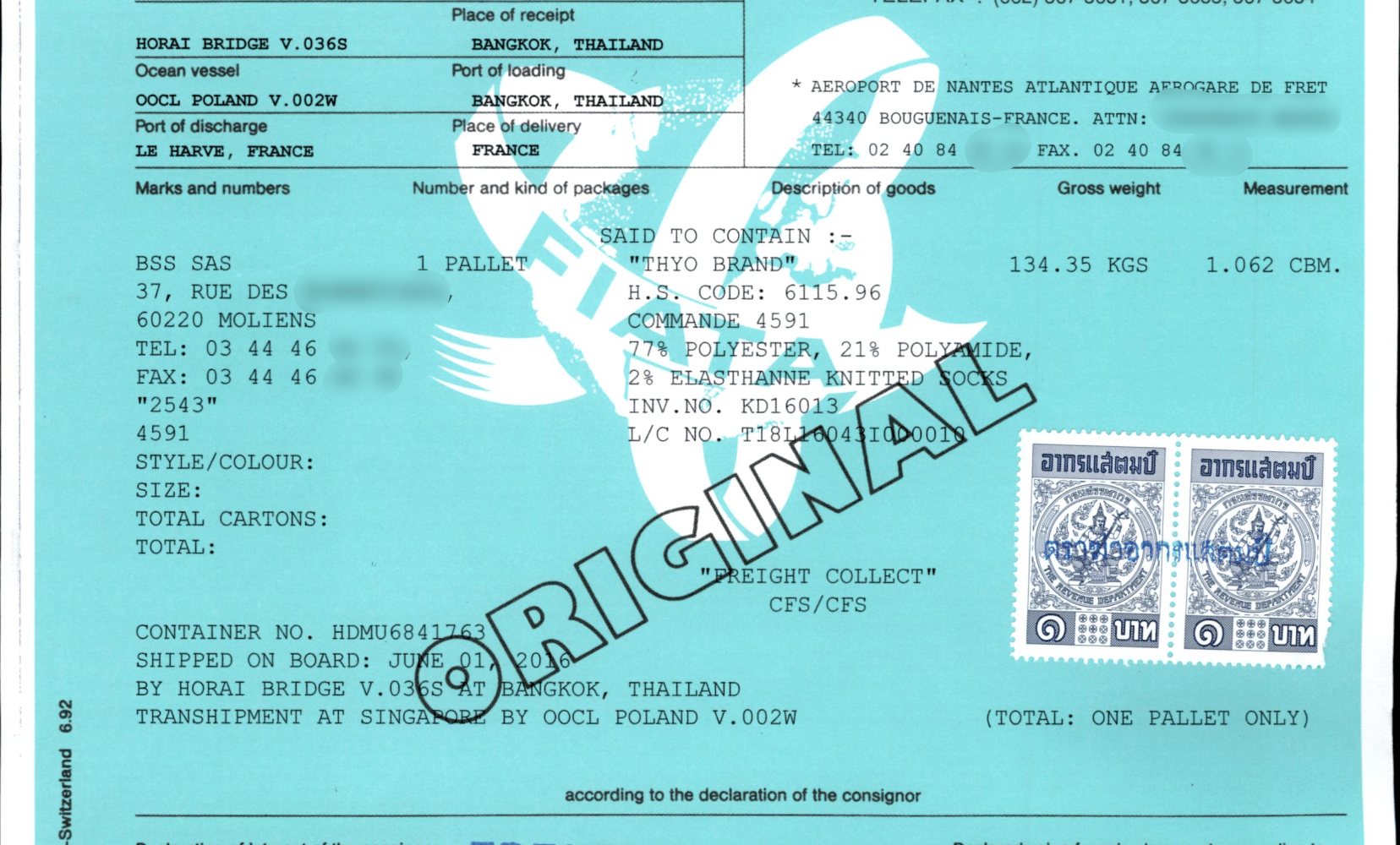}  \\
    \includegraphics[width=80mm, height=50mm]{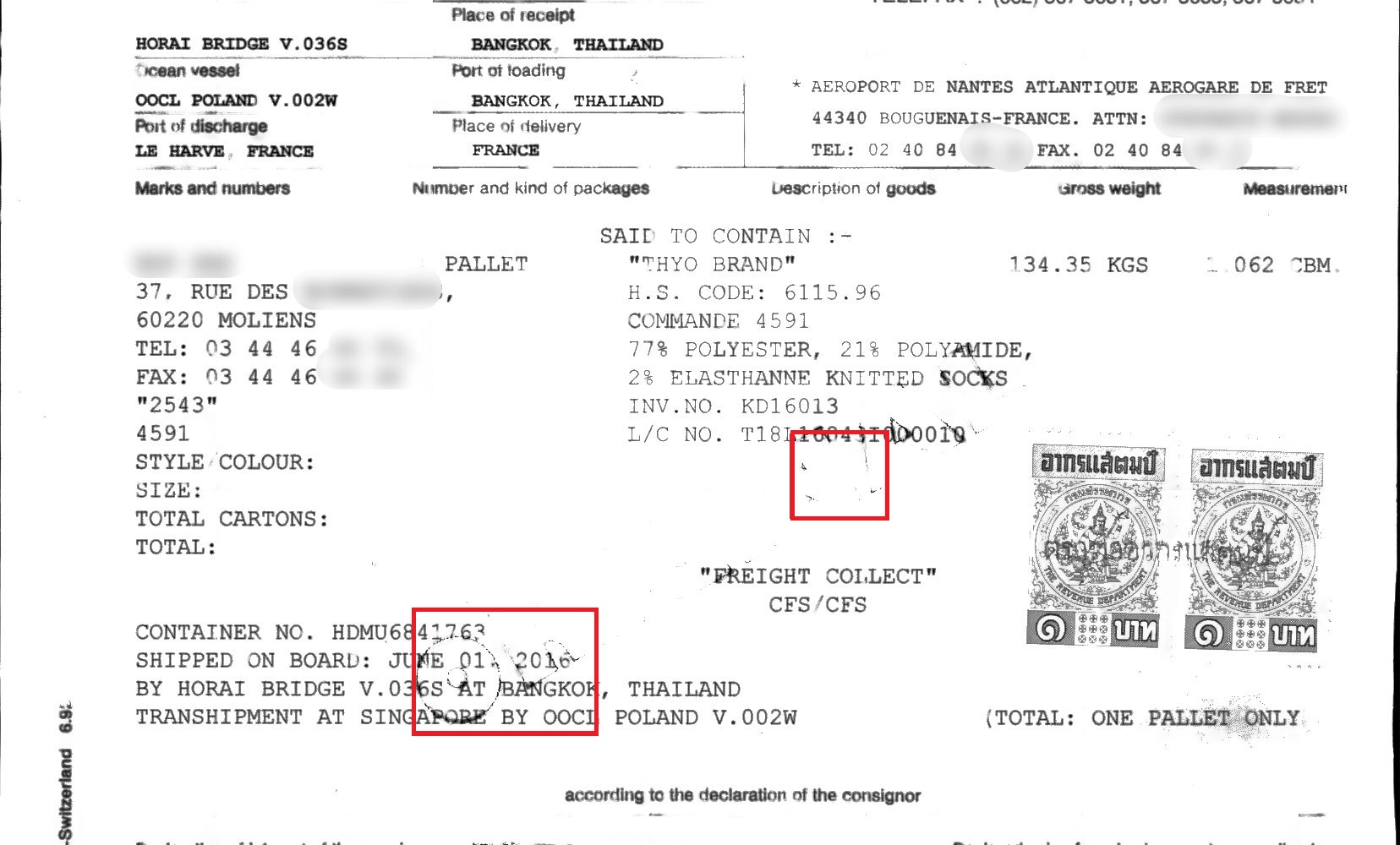}  \\
    
\end{tabular}
\caption{Qualitative results for dense watermark removal. Above, a section from watermarked invoice. Below, it’s enhanced version. Some parts of the text in the invoice was blurred due to privacy constraints. Because of different domains, synthetic vs real, we can see that some tiny parts of the watermark were not completely removed (red boxes).}
\label{stamprem}
\end{figure}

\begin{figure*}[h]
\centering
\includegraphics[width=170mm, height=188mm]{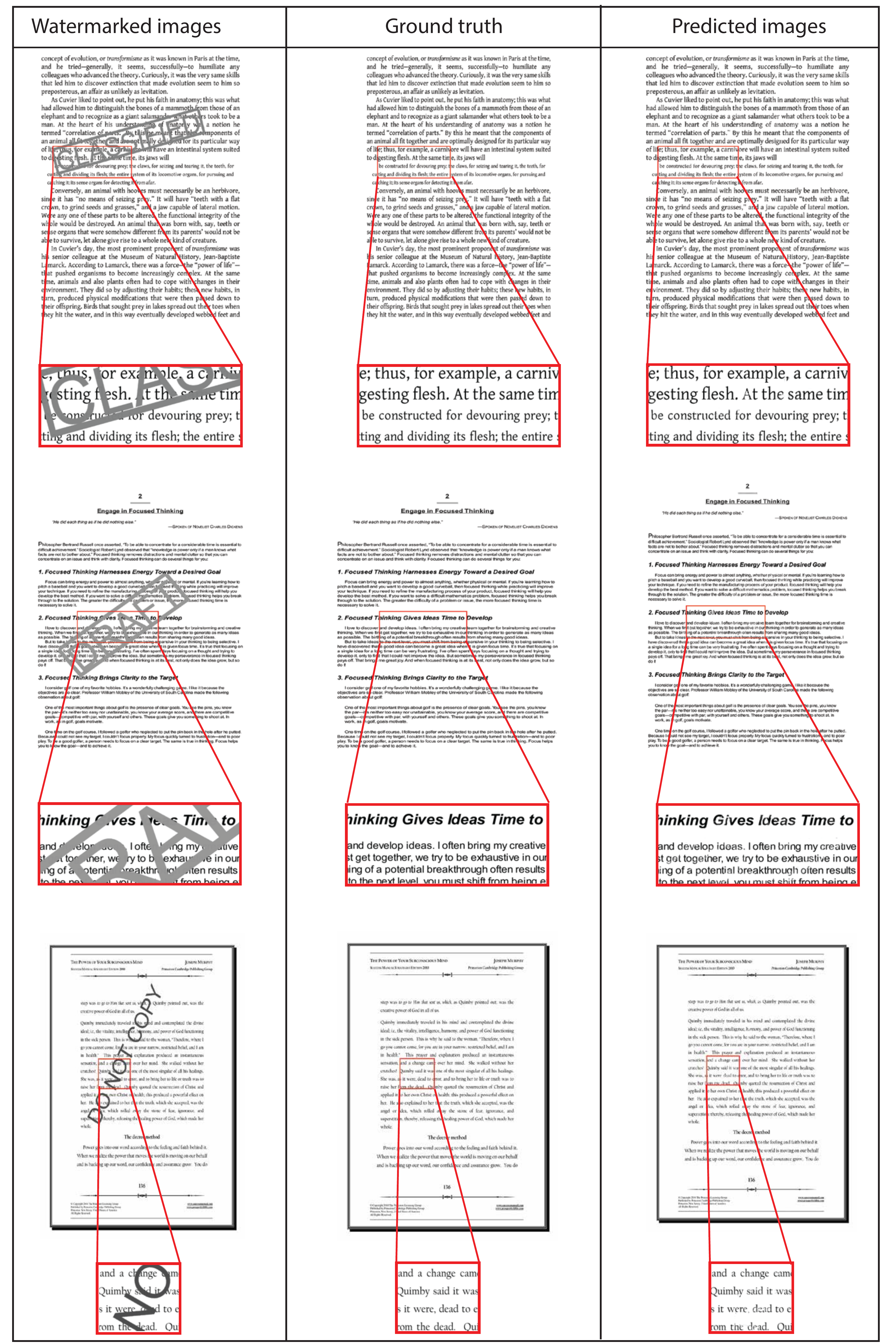}
\caption{Watermark removal by DE-GAN} \label{figwatermark}
\end{figure*}

Despite  that the watermarks used in our study were very dense and we believe that removing them is harder than the related approaches presented in Table \ref{tab3}.  Our approach surpasses, by far, those in natural images. Fig. \ref{figwatermark} shows some examples of watermark removal by DE-GAN, the produced images are preserving the text quality while removing the foreground watermarks. In addition, since the presented watermarked documents were  synthetically made, it was interesting to  apply DE-GAN to remove watermarks from a naturally degraded document. Fig. \ref{stamprem} shows that DE-GAN  successfully  removes a  dense watermark from a document paper. As you can see, the watermark is completely removed, and the reader or the OCR system can easily read the enhanced document compared to the degraded one.

\begin{table}[ht]
\centering
\caption{Results of watermark removal}\label{tab3}
\begin{tabular}{|l|l|l|}
\hline
Model  & PSNR &   SSIM\\
\hline
 Dekel et al. \cite{art11} &   36.02  &   0.924\\
\hline
Wu et al. \cite{art12} &   23.37 &   0.884\\
\hline
Cheng et al. \cite{art14} &   30.86 &   0.914\\
\hline
\textbf{DE-GAN}  &   \textbf{40.98} &   \textbf{0.998}\\
\hline
\end{tabular}
\end{table}

%%%%%%%%%%%%%%%%%%%%%%%%%%%%%%%%%%%%%%%%%%%%%%%%%%%%%%%%%%%%%%%%%%%%%%%%%%%%%%%%%%%%%%%%%%%%%%%%%%%%%%%%%%%%%%%%%%%%%%%%%%%%%%%%%%%%%%

\subsection{Comparison with other GAN models}

\begin{figure}[ht]
\centering
\begin{tabular}{c}
  \includegraphics[width=70mm, height=25mm]{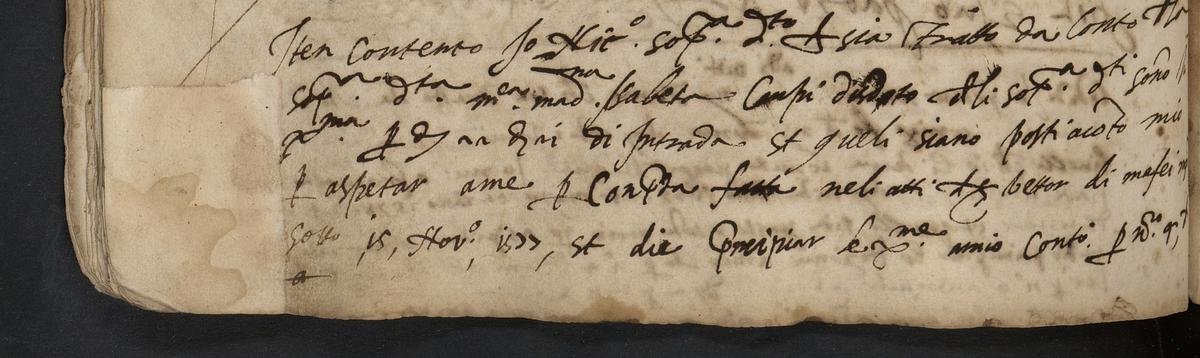}\\
Original \\[6pt]
  \includegraphics[width=70mm, height=25mm]{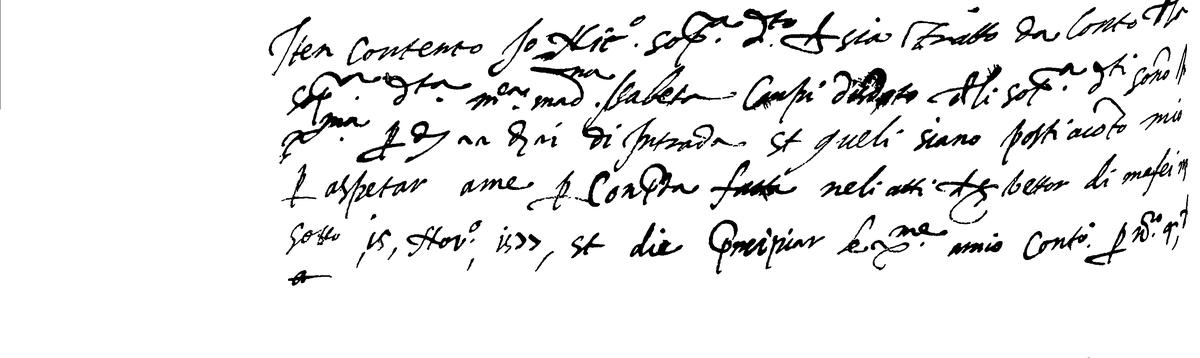}\\
Ground truth \\[6pt]
  \includegraphics[width=70mm, height=25mm]{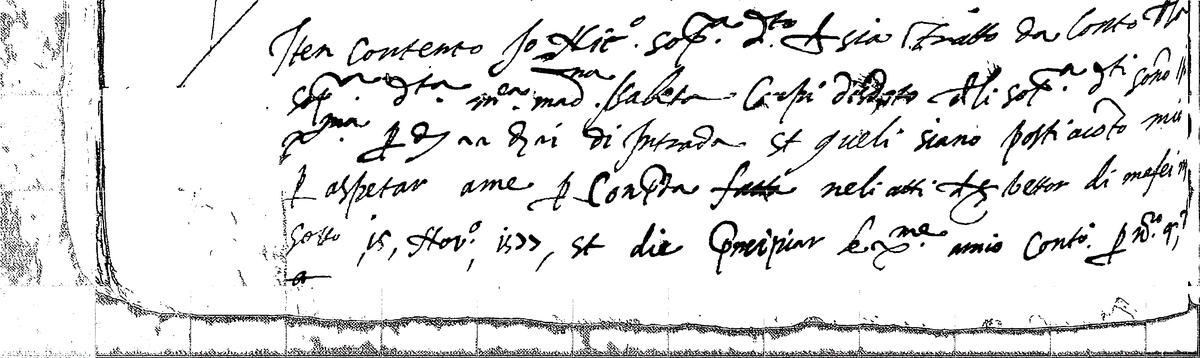}\\
CycleGan \\[6pt]
  \includegraphics[width=70mm, height=25mm]{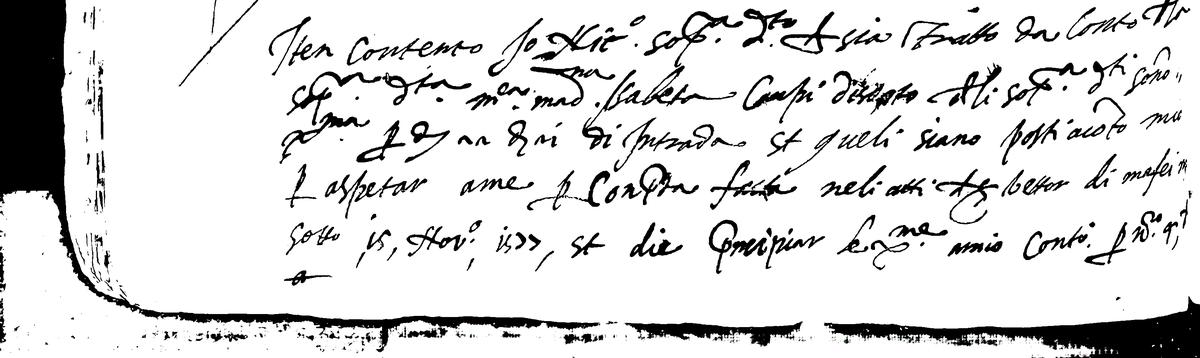}\\
Pix2pix-HD \\[6pt]
  \includegraphics[width=70mm, height=25mm]{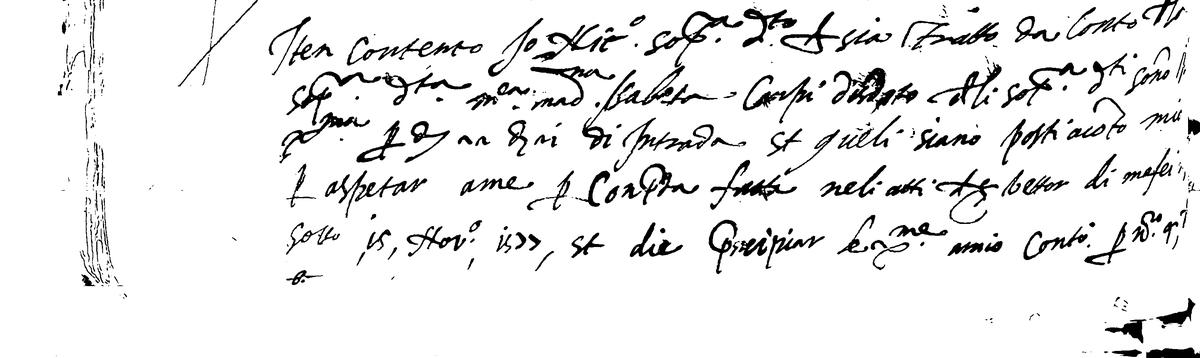}\\
DE-GAN \\[6pt]
\end{tabular}
\caption{Qualitative binarization results  produced by different models of the sample (9) from H-DIBCO 2018 dataset}
\label{figcompgan}
\end{figure}

As it is a fact that our model is inspired from the pix2pix model \cite{art2} (we are using a deeper generator and a different additional loss), it would be useful if we tried some other similar  models that are based on GANs and dedicated to the same image-to-image translation problem. For this aim, cycleGAN \cite{art40} and pix2pix-HD \cite{art62} models are considered for the comparison. We evaluate these models on H-DIBCO 2018 dataset \cite{art58} with the same conditions and data used to train the DE-GAN. 
The quantitative and qualitative obtained results are presented in Table \ref{tabgans} and Fig. \ref{figcompgan}, respectively. Experimental results shows the superiority of DE-GAN compared to cycleGAN and pix2pix-HD in achieving higher PSNR, F-measure and Fps and a lower DRD. 
We note that the unsupervised training capabilities of CycleGAN are quite useful since paired data is harder to find in document enhancement applications. For pix2pix-HD, the results are promising, since the training samples that we used for training were few (the number of DIBCO samples is small if we split them to patches with size $512\times1024$, that's why we used some flips of images to augment the data). With more data, we believe that pix2pix-HD could perform much better.

\begin{table}[ht]
\centering
\caption{Results of image binarization for DIBCO 2018 Database}\label{tabgans}
\begin{tabular}{|l|l|l|l|l|}
\hline
Model  & PSNR & F-measure & F$_{ps}$&DRD\\
\hline
{cycleGAN}  & 11.00 & 56.33 &58.07& 30.07\\
\hline
{pix2pix-HD} & 14.42 & 72.79 &76.28& 15.13\\
\hline
\textbf{DE-GAN}   & \textbf{16.16} & \textbf{77.59} &\textbf{85.74}& \textbf{7.93}\\
\hline
\end{tabular}
\end{table}

\subsection{Document deblurring}
\begin{figure}[ht]
\centering
\begin{tabular}{ccccc}

 \includegraphics[width=12mm, height=16mm]{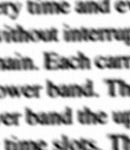} &   \includegraphics[width=12mm, height=16mm]{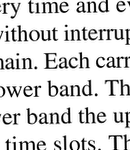}  &   \includegraphics[width=12mm, height=16mm]{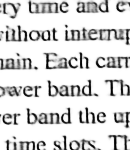} &
  \includegraphics[width=12mm, height=16mm]{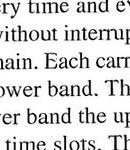}  &   \includegraphics[width=12mm, height=16mm]{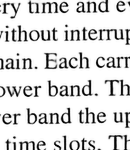} \\[6pt]
  
  \includegraphics[width=12mm, height=16mm]{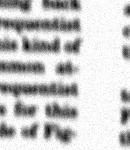} &   \includegraphics[width=12mm, height=16mm]{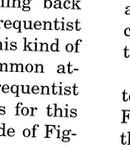}  &   \includegraphics[width=12mm, height=16mm]{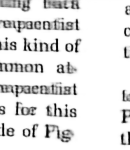} &
  \includegraphics[width=12mm, height=16mm]{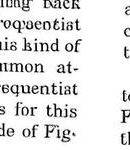}  &   \includegraphics[width=12mm, height=16mm]{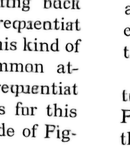} \\[6pt]
  
  \includegraphics[width=12mm, height=16mm]{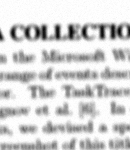} &   \includegraphics[width=12mm, height=16mm]{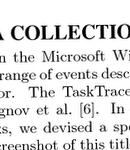}  &   \includegraphics[width=12mm, height=16mm]{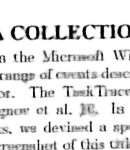} &
  \includegraphics[width=12mm, height=16mm]{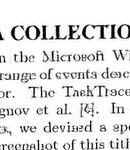}  &   \includegraphics[width=12mm, height=16mm]{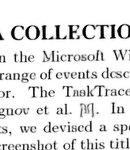} \\[6pt]
  
  \includegraphics[width=12mm, height=16mm]{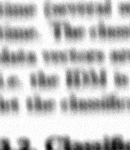} &   \includegraphics[width=12mm, height=16mm]{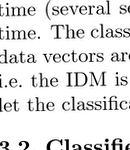}  &   \includegraphics[width=12mm, height=16mm]{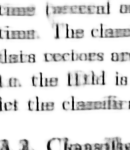} &
  \includegraphics[width=12mm, height=16mm]{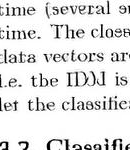}  &   \includegraphics[width=12mm, height=16mm]{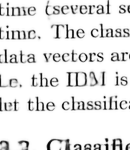} \\
  Original& GT& \vtop{\hbox{\strut CNN}\hbox{\strut \cite{art60}}} &\vtop{\hbox{\strut pix2pix-HD}\hbox{\strut \cite{art62}}}& DE-GAN
  
\end{tabular}
\caption{Qualitative deblurring results of some patches produced by different methods}
\label{figblur}
\end{figure}

The DE-GAN model presented in this paper is able to outperform many state-of-the-art approaches in different problems like binarization, denoising and watermark removal. To experimentally prove the efficiency and the flexibility of the proposed method, we evaluate it on a more challenging scenario, which is document deblurring. We use $4000$ patches from the dataset developed in \cite{art60} to train our model, and $932$ patches for testing. 
Noting that, in \cite{art60} a convolutional neural network architecture is proposed to address the problem. Thus, we will compare the results with this CNN and pix2pix-HD models trained on this selected data. The obtained results are presented in Table \ref{tableblur}. We can see that GAN's models surpasses the CNN. This is much clear in the qualitative results of some patches presented in   Fig \ref{figblur}. We can also see that DE-GAN gives similar results to pix2pix-HD, however, it is more accurate for predicting some characters. For example, in the second patch row, third line, the word "kind" is correctly predicted by DE-GAN but it is predicted as "bind" by pix2pix-HD. We note that the used dataset is composed of 300x300px image patches, which can explain why pix2pix-HD does not give a better performance (it works generally with larger input patch with a size of 512x1024, or 1024x2048).  

\begin{table}[ht]
\centering
\caption{The obtained results of document deblurring}\label{tableblur}
\begin{tabular}{|l|l|}
\hline
Method & PSNR\\
\hline
  CNN \cite{art60} & 19.36\\
\hline
pix2pix-HD \cite{art62} & 19.89\\
\hline
\textbf{DE-GAN}  & \textbf{20.37}\\
\hline
\end{tabular}
\end{table}

\subsection{OCR evaluation}

\begin{figure}[ht]
\centering
\begin{tabular}{|c|}
\hline\\
    \includegraphics[width=70mm, height=4mm]{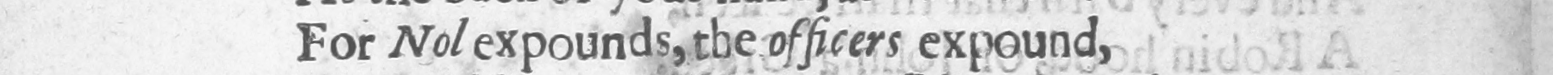}  \\
    For Nol expoundsytbe officers expound,  \\[0.5 cm]
    % \hline
  
    \includegraphics[width=70mm, height=4mm]{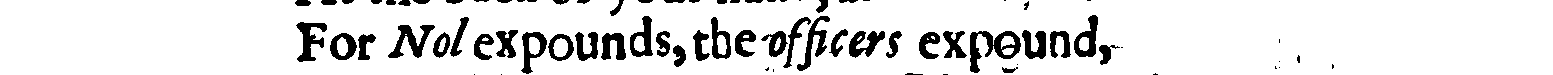}  \\
    For Nol expounds, the officers expound,  \\
  \hline\\
%     \includegraphics[width=70mm, height=4mm]{images/l2.png}  \\
%     Augustin $\mid$ a feuilleter $\mid$ les cartons, de  grayur,  \\[0.5 cm]
% %   \hline
%      \includegraphics[width=70mm, height=4mm]{images/l2e.png}  \\
%   Augustin a feuilleter les cartons de gravu-  \\
%   \hline\\
     \includegraphics[width=70mm, height=4mm]{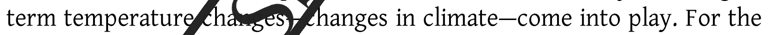}  \\
     \vtop{\hbox{\strut term temperaturef ham Yhanges }\hbox{\strut in climate-come into play. For the }} \\[0.5 cm]
    
%   \hline
   \includegraphics[width=70mm, height=4mm]{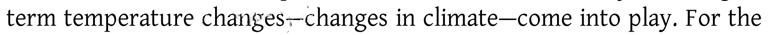}  \\
    \vtop{\hbox{\strut term temperature changes-changes }\hbox{\strut in climate-come into play. For the }} \\
\hline
\end{tabular}
\caption{Qualitative results for Tesseract recognition of some text lines}
\label{ocrtab}
\end{figure}

After the quantitatively and qualitatively evaluation of the resulted enhanced images presented previously, we compare in what follows the performance of OCR  on degraded and enhanced documents.  For this aim, we took a set of $4$ images ($2$ degraded ones from DIBCO datasets, and $2$ images with a dense watermark from our dataset).Then, we used Tesseract OCR \cite{art59} to recognize those images and their enhanced versions with DE-GAN. We found that the proposed enhancement method boosts the baseline OCR performance by a large margin, the character error rate is decreased from $0.37$ for the degraded documents to $0.01$ for the enhanced ones. Fig. \ref{ocrtab} shows a tiny example of this process. In each row, you can find a line of a degraded document image  and the text produced by the OCR system, then its enhanced version followed the OCR text.

\section{Conclusion}
In this paper we proposed  a Document Enhancement Generative Adversarial Network named DE-GAN to restore severely degraded document images. 
DE-GAN is a modified version of the pix2pix with a deeper generator and a different additional loss (adversarial + log) to generate an enhanced document given the degraded version. To the best of our knowledge, this is the first application of GANs for studying document enhancement problem.  Moreover, we present a new problem in document enhancement that is dense watermark (or stamps) removal, hoping that it takes the attention of document analysis community.  Extensive experiments show that DE-GAN achieved interesting results in different document enhancement tasks that outperform the fully convolutional networks, cycleGAN and pix2pix-HD models. Furthermore, we achieve improved results compared to many recent state-of-the-art methods on benchmarking datasets like DIBCO 2013, DIBCO 2017 and H-DIBCO 2018. 

We showed  that the proposed enhancement method boosts the baseline OCR performance by a large margin. Hence,  as an immediate future work, we plan to add the OCR evaluation in the discriminator part. Thus, we can give the discriminator the ability of reading the text to decide if it is real or fake, which will force it to generate more readable images. We intend, also,  to test the performance of the DE-GAN on mobile captured documents which present many problems like shadow, real blur, low resolution, distortion, etc.

\section*{Acknowledgement}
This work has been partially supported by the Swedish Research Council (grant 2018-06074, DECRYPT), the Spanish project RTI2018-095645-B-C21 and the CERCA Program / Generalitat de Catalunya..

\ifCLASSOPTIONcaptionsoff
  \newpage
\fi



%\bibliographystyle{IEEEtran}

\bibliographystyle{ieeetr}
\bibliography{bibl}

\begin{thebibliography}{10}

\bibitem{art31}
K.~Chellapilla, S.~Puri, and P.~Simard, ``High performance convolutional neural
  networks for document processing,'' in {\em International Workshop on
  Frontiers in Handwriting Recognition}, 2006.

\bibitem{art32}
S.~Schreiber, S.~Agne, I.~Wolf, A.~Dengel, and S.~Ahmed, ``Deepdesrt: Deep
  learning for detection and structure recognition of tables in document
  images,'' in {\em ICDAR}, 2017.

\bibitem{art19}
F.~Zamora-Martinez, S.~España-Boquera, and M.~J. Castro-Bleda,
  ``Behaviour-based clustering of neural networks applied to document
  enhancement,'' in {\em Computational and Ambient Intelligence}, pp.~144--151,
  2007.

\bibitem{art30}
S.~Bako, S.~Darabi, E.~Shechtman, J.~Wang, K.~Sunkavalli, and P.~Sen,
  ``Removing shadows from images,'' in {\em ACCV}, 2016.

\bibitem{art37}
X.~Chen, X.~He, J.~Yang, and Q.~Wu, ``An effective document image deblurring
  algorithm,'' in {\em CVPR}, 2011.

\bibitem{art1}
X.-J. Mao, C.~Shen, and Y.-B. Yang, ``Image restoration using very deep
  convolutional encoder-decoder networks with symmetric skip connections,'' in
  {\em NIPS}, 2016.

\bibitem{art46}
C.~Dong, C.~C. Loy, K.~He, and X.~Tang, ``Image super-resolution using deep
  convolutional networks,'' {\em IEEE transactions on pattern analysis and
  machine intelligence}, vol.~38, pp.~295--307, 2016.

\bibitem{art66}
D.~P. Kingma and M.~Welling, ``Auto-encoding variational bayes,'' {\em arXiv
  preprint arXiv}, 2013.

\bibitem{art2}
P.~Isola, J.-Y. Zhu, T.~Zhou, and A.~A. Efros, ``Image-to-image translation
  with conditional adversarial networks,'' in {\em CVPR}, 2017.

\bibitem{art18}
T.~Karras, T.~Aila, S.~Laine, and J.~Lehtinen, ``Progressive growing of gans
  for improved quality, stability, and variation,'' {\em arXiv preprint arXiv},
  2017.

\bibitem{art13}
I.~Goodfellow, J.~Pouget-Abadie, M.~Mirza, B.~Xu, D.~Warde-Farley, S.~Ozair,
  A.~Courville, and Y.~Bengio, ``Generative adversarial nets,'' in {\em
  Advances in neural information processing systems}, pp.~2672--2680, 2014.

\bibitem{art55}
A.~K. Bhunia, A.~K. Bhunia, P.~Banerjee, A.~Konwer, A.~Bhowmick, P.~P. Roy, and
  U.~Pal, ``Word level font-to-font image translation using convolutional
  recurrent generative adversarial networks,'' {\em arXiv preprint arXiv},
  2018.

\bibitem{art33}
A.~Ghosh, B.~Bhattacharya, and S.~B.~R. Chowdhury, ``Handwriting profiling
  using generative adversarial networks,'' in {\em AAAI Conference on
  Artificial Intelligence}, 2017.

\bibitem{art34}
A.~Konwer, A.~K. Bhunia, A.~Bhowmick, A.~K. Bhunia, P.~Banerjee, P.~P. Roy, and
  U.~Pal, ``Staff line removal using generative adversarial networks,'' in {\em
  ICPR}, pp.~1103--1108, 2018.

\bibitem{art35}
N.~Kligler, S.~Katz, and A.~Tal, ``Document enhancement using visibility
  detection,'' in {\em CVPR}, pp.~2374--2382, 2018.

\bibitem{art36}
R.~K. Pandey and A.~G. Ramakrishnan, ``Language independent single document
  image super-resolution using cnn for improved recognition,'' {\em arXiv
  preprint arXiv}, 2017.

\bibitem{art45}
C.~L. Tan, L.~Zhang, Z.~Zhang, and T.~Xia, ``Restoring warped document images
  through 3d shape modeling,'' {\em IEEE transactions on pattern analysis and
  machine intelligence}, vol.~28, pp.~195--208, 2006.

\bibitem{art3}
N.~Otsu, ``A threshold selection method from gray-level histograms,'' {\em IEEE
  transactions on systems}, vol.~9, no.~1, pp.~62--66, 1979.

\bibitem{art42}
J.~Sauvola and M.~Pietik, ``Adaptive document image binarization,'' {\em
  Pattern recognition}, vol.~33, pp.~225--236, 2000.

\bibitem{art43}
W.~Niblack, {\em An introduction to digital image processing}.
\newblock Strandberg Publishing Company Birkeroed, 1985.

\bibitem{art4}
N.~Phansalkart, S.~More, A.~Sabale, and M.~Joshi, ``Adaptive local thresholding
  for detection of nuclei in diversity stained cytology images,'' in {\em
  ICCSP}, 2011.

\bibitem{art5}
G.~chutani, T.~Patnaik, and V.~Dwivedi, ``An improved approach for automatic
  denoising and binarization of degraded document images based on region
  localization,'' in {\em ICACCI}, 2015.

\bibitem{art47}
M.~Cheriet, J.~N. Said, and C.~Y. Suen, ``A recursive thresholding technique
  for image segmentation,'' {\em IEEE transactions on pattern analysis and
  machine intelligence}, vol.~7, pp.~918--921, 1998.

\bibitem{art49}
T.~Lelore and F.~Bouchara, ``Fair: A fast algorithm for document image
  restoration,'' {\em IEEE transactions on pattern analysis and machine
  intelligence}, vol.~35, pp.~2039--2048, 2013.

\bibitem{art51}
M.~Annabestani and M.~Saadatmand-Tarzjan, ``A new threshold selection method
  based on fuzzy expert systems for separating text from the background of
  document images,'' {\em Iranian journal of science and technology,
  transactions of electrical engineering}, pp.~1--13, 2018.

\bibitem{art9}
Chien-HsingChou, Wen-HsiungLin, and FuChang, ``A binarization method with
  learning-built rules for document images produced by cameras,'' {\em Pattern
  Recognition}, vol.~43, pp.~1518--1530, 2010.

\bibitem{art52}
W.~Xiong, J.~Xu, Z.~Xiong, J.~Wang, and M.~Liu, ``Degraded historical document
  image binarization using local features and support vector machine (svm),''
  {\em Optik}, vol.~164, pp.~218--223, 2018.

\bibitem{art48}
R.~F. Moghaddam and M.~Cheriet, ``A variational approach to degraded document
  enhancement,'' {\em IEEE transactions on pattern analysis and machine
  intelligence}, vol.~32, pp.~1347--1361, 2010.

\bibitem{art6}
R.~Hedjam, M.~Cheriet, and M.~Kalacska, ``Constrained energy maximization and
  self-referencing method for invisible ink detection from multispectral
  historical document images,'' in {\em ICPR}, 2014.

\bibitem{art56}
S.~Milyaev, O.~Barinova, T.~Novikova, P.~Kohli, and V.~Lempitsky, ``Fast and
  accurate scene text understanding with image binarization and off-the-shelf
  ocr,'' {\em International Journal on Document Analysis and Recognition},
  2015.

\bibitem{art63}
W.~Xiong, X.~Jia, J.~Xu, Z.~Xiong, M.~Liu, and J.~Wang, ``Historical document
  image binarization using background estimation and energy minimization,'' in
  {\em ICPR}, 2018.

\bibitem{art54}
G.~Meng, K.~Yuan, Y.~Wu, S.~Xiang, and C.~Pan, ``Deep networks for degraded
  document image binarization through pyramid reconstruction,'' in {\em ICDAR},
  pp.~2379--2140, 2017.

\bibitem{art64}
J.~Calvo-Zaragoza and A.-J. Gallego, ``A selectional auto-encoder approach for
  document image binarization,'' {\em Pattern Recognition}, vol.~86,
  pp.~37--47, 2019.

\bibitem{art65}
K.~G. Lore, A.~Akintayo, and S.~Sarkar, ``Llnet: A deep autoencoder approach to
  natural low-light image enhancement,'' {\em Pattern Recognition}, vol.~61,
  pp.~650--662, 2017.

\bibitem{art50}
M.~Z. Afzal, J.~Pastor-Pellicer, F.~Shafait, T.~M. Breuel, A.~Dengel, and
  M.~Liwicki, ``Document image binarization using lstm: A sequence learning
  approach,'' in {\em HIP@ICDAR}, pp.~79--84, 2015.

\bibitem{art8}
C.~Tensmeyer and T.~Martinez, ``Document image binarization with fully
  convolutional neural networks,'' in {\em ICDAR}, pp.~99--104, 2017.

\bibitem{art53}
F.~Westphal, N.~Lavesson, and H.~Grahn, ``Document image binarization using
  recurrent neural networks,'' in {\em IAPR}, pp.~263--268, 2018.

\bibitem{art7}
Q.~N. Vo, S.~H. Kim, H.~J. Yang, and G.~Lee, ``Binarization of degraded
  document images based on hierarchical deep supervised network,'' {\em Pattern
  Recognition}, vol.~74, pp.~568--586, 2018.

\bibitem{art10}
C.~Xu, Y.~Lu, and Y.~Zhou, ``An automatic visible watermark removal technique
  using image inpainting algorithms,'' in {\em ICSAI}, pp.~1152--1157, 2017.

\bibitem{art11}
T.~Dekel, M.~Rubinstein, and W.~T.~F. Ce~Liu, ``On the effectiveness of visible
  watermarks,'' in {\em CVPR}, pp.~2146--2154, 2017.

\bibitem{art12}
J.~Wu, H.~Shi, S.~Zhang, Z.~Lei, Y.~Yang, and S.~Z. Li, ``De-mark gan: Removing
  dense watermark with generative adversarial network,'' in {\em CVPR},
  pp.~2146--2154, 2018.

\bibitem{art14}
D.~Cheng, X.~Li, W.-H. Li, C.~Lu, F.~Li, H.~Zhao, and W.-S. Zheng,
  ``Large-scale visible watermark detection and removal with deep convolutional
  networks,'' in {\em PRCV}, pp.~27--40, 2018.

\bibitem{art38}
P.~Luc, C.~Couprie, S.~Chintala, and J.~Verbeek, ``Semantic segmentation using
  adversarial networks,'' {\em arXiv preprint arXiv}, 2016.

\bibitem{art39}
C.~Ledig, L.~Theis, F.~Huszár, J.~Caballero, A.~Cunningham, A.~Acosta,
  A.~Aitken, A.~Tejani, J.~Totz, Z.~Wang, and W.~Shi, ``Photo-realistic single
  image super-resolution using a generative adversarial network,'' in {\em
  CVPR}, 2016.

\bibitem{art40}
J.-Y. Zhu, T.~Park, P.~Isola, and A.~A. Efros, ``Unpaired image-to-image
  translation using cycle-consistent adversarial networks,'' {\em arXiv
  preprint arXiv}, 2017.

\bibitem{art62}
T.-C. Wang, M.-Y. Liu, J.-Y. Zhu, A.~Tao, J.~Kautz, and B.~Catanzaro,
  ``High-resolution image synthesis and semantic manipulation with conditional
  gans,'' in {\em CVPR}, 2018.

\bibitem{art41}
Z.~Yi, H.~Zhang, P.~Tan, , and M.~Gong, ``Dualgan: unsupervised dual learning
  for image-to-image translation,'' {\em arXiv preprint arXiv}, 2018.

\bibitem{art17}
O.~Ronneberger, P.~Fischer, and T.~Brox, ``U-net: Convolutional networks for
  biomedical image segmentation,'' in {\em MICCAI}, pp.~234--241, 2015.

\bibitem{art44}
E.~Shelhamer, J.~Long, and T.~Darrell, ``Fully convolutional networks for
  semantic segmentation,'' {\em IEEE transactions on pattern analysis and
  machine intelligence}, vol.~39, pp.~640--651, 2016.

\bibitem{art15}
J.~Xie, L.~Xu, and E.~Chen, ``Image denoising and inpainting with deep neural
  networks,'' in {\em NIPS}, pp.~350--358, 2012.

\bibitem{art16}
V.~Badrinarayanan, A.~Kendall, and R.~Cipolla, ``Segnet: A deep convolutional
  encoder-decoder architecture for scene segmentation,'' {\em IEEE transactions
  on pattern analysis and machine intelligence}, vol.~39, pp.~2481--2495, 2017.

\bibitem{art20}
I.~Pratikakis, B.~Gatos, , and K.~Ntirogiannis, ``Icdar 2013 document image
  binarization contest (dibco 2013),'' in {\em ICDAR}, pp.~1471--1476, 2013.

\bibitem{art22}
B.~Gatos, K.~Ntirogiannis, and I.~Pratikakis, ``Icdar 2009 document image
  binarization contest (dibco 2009),'' in {\em ICDAR}, pp.~1375--1382, 2009.

\bibitem{art23}
I.~Pratikakis, B.~Gatos, and K.~Ntirogiannis, ``Icdar 2011 document image
  binarization contest (dibco 2011),'' in {\em ICDAR}, pp.~1506--1510, 2011.

\bibitem{art24}
I.~Pratikakis, B.~Gatos, and K.~Ntirogiannis, ``Icfhr 2012 competition on
  handwritten document image binarization (h-dibco 2012),'' in {\em ICFHR},
  pp.~817--822, 2012.

\bibitem{art21}
K.~Ntirogiannis, B.~Gatos, and I.~Pratikakis, ``Icfhr2014 competition on
  handwritten document image binarization (h-dibco 2014),'' in {\em ICFHR},
  pp.~809--813, 2014.

\bibitem{art25}
I.~Pratikakis, K.~Zagoris, G.~Barlas, and B.~Gatos, ``Icfhr 2016 competition on
  handwritten document image binarization (h-dibco 2016),'' in {\em ICFHR},
  pp.~2167--6445, 2016.

\bibitem{art26}
I.~Pratikakis, K.~Zagoris, G.~Barlas, and B.~Gato, ``Icdar2017 competition on
  document image binarization (dibco 2017),'' in {\em ICDAR}, pp.~2379--2140,
  2017.

\bibitem{art29}
B.~Gatos, I.~Pratikakis, , and S.~Perantonis, ``An adaptive binarization
  technique for low quality historical documents,'' in {\em
  InternationalWorkshop on Document Analysis Systems}, pp.~102--113, 2014.

\bibitem{art28}
B.~Su, S.~Lu, and C.~Tan, ``Robust document image binarization technique for
  degraded document images,'' {\em IEEE transactions on image processing},
  vol.~22, pp.~1408--1417, 2013.

\bibitem{art27}
N.~Howe, ``Document binarization with automatic parameter tuning,'' {\em
  International Journal on Document Analysis and Recognition (IJDAR)}, vol.~16,
  pp.~247--258, 2013.

\bibitem{art58}
I.~Pratikakis1, K.~Zagoris1, P.~Kaddas, and B.~Gatos, ``Icfhr2018 competition
  on handwritten document image binarization (h-dibco 2018),'' in {\em ICFHR},
  pp.~489--493, 2018.

\bibitem{art60}
M.~Hradiš, J.~Kotera, P.~Zemcík, and F.~Šroubek, ``Convolutional neural
  networks for direct text deblurring,'' in {\em BMVC}, 2015.

\bibitem{art59}
R.~Smith, ``An overview of the tesseract ocr engine,'' in {\em ICDAR},
  pp.~629--633, 2007.

\end{thebibliography}
% \begin{thebibliography}{bibl}


% \end{thebibliography}


\begin{IEEEbiography}[{\includegraphics[width=1in,height=1.25in,clip,keepaspectratio]{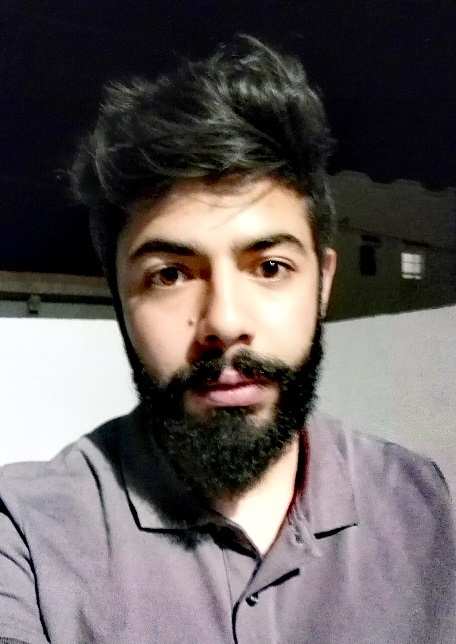}}]{Mohamed Ali Souibgui} Received his bachelor and master degrees in computer science from the University of Monastir, Tunisia, in 2015 and 2018, respectively. He is currently working toward the PhD degree in the Computer Vision Center, Autonomous University of
Barcelona, Spain. His research interests include machine learning, computer vision and document analysis.

%or if you just want to reserve a space for a photo:
\end{IEEEbiography}

\begin{IEEEbiography}[{\includegraphics[width=1in,height=1.25in,clip,keepaspectratio]{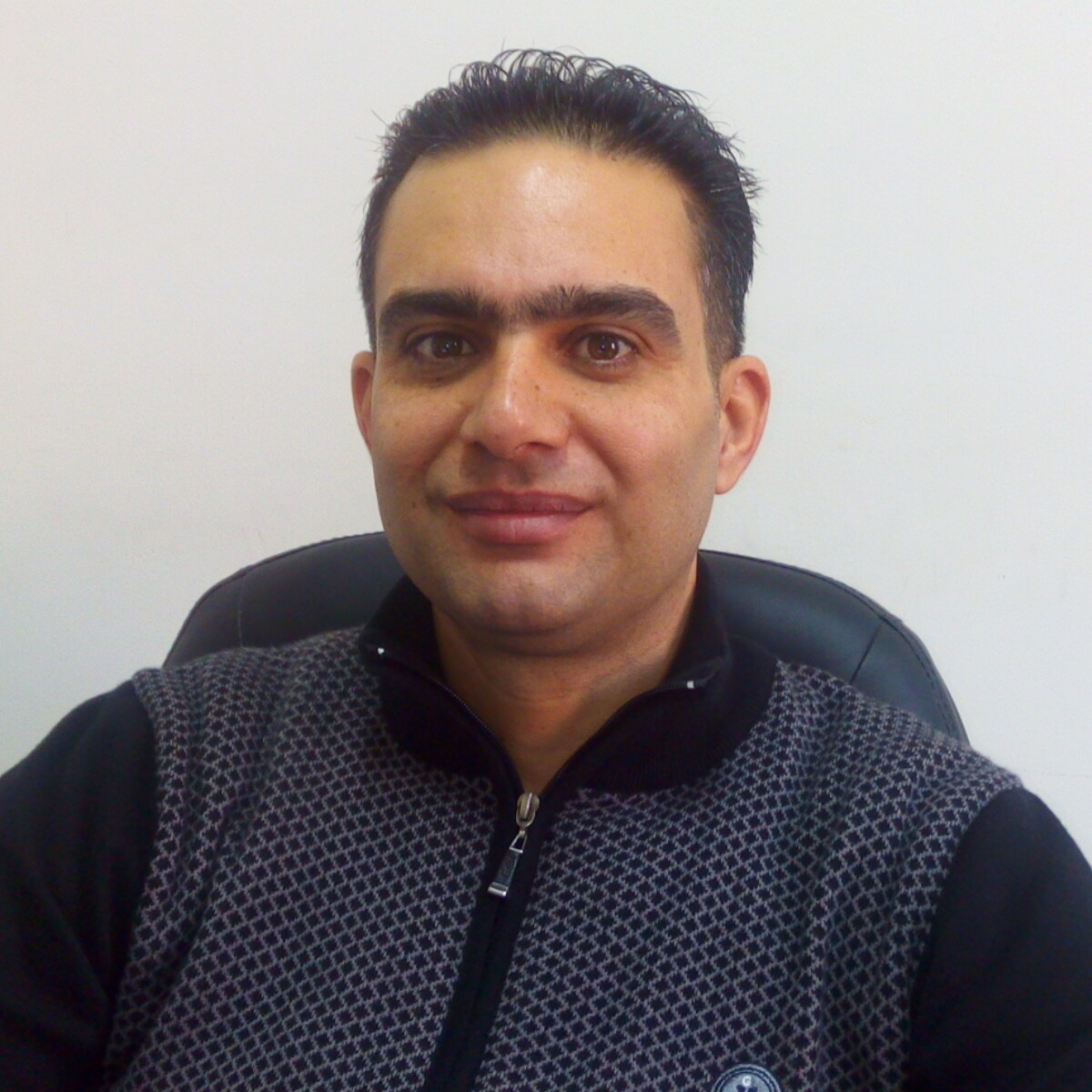}}]{Yousri Kessentini} is graduated in Computer Science engineering from the National Engineering School of Sfax (ENIS) in 2003 and received his Ph.D. degree in the field of pattern recognition from the University of Rouen, France in 2009. He was postdoctoral researcher at ITESOFT company and LITIS laboratory from 2011 to 2013. Currently he is Assistant Professor at  CRNS and  the head of the DeepVision research team. His main research areas concern deep learning, document processing and recognition, data fusion, and computer vision. He is certified as an official instructor and ambassador from the NVIDIA Deep Learning Institute. He has coordinate several research projects in partnership with industry and  the author and co-author of several papers. 
%or if you just want to reserve a space for a photo:
\end{IEEEbiography}


%\newpage



%\vfill


\end{document}